%% file: CWNN_Arxiv_20250712.tex
\documentclass[journal]{IEEEtran}

\usepackage{amsmath,amsopn,amssymb}
\usepackage{cite}

\usepackage{graphicx}
\usepackage{epstopdf}
\usepackage{subfigure}
\usepackage{stfloats}

\usepackage[colorlinks,linkcolor=red,anchorcolor=blue,citecolor=blue]{hyperref}
\usepackage{color}

\usepackage[mathscr]{eucal}
\usepackage{threeparttable}
\usepackage{lastpage}
\usepackage{bm}
\usepackage{lineno}
\usepackage{float}
\usepackage{multirow}
\usepackage[linesnumbered,ruled,lined]{algorithm2e}

\allowdisplaybreaks

\newtheorem{lemma}{Lemma}
\newtheorem{theorem}{Theorem}
\newtheorem{corollary}{Corollary}

\newtheorem{remark}{Remark}

\newtheorem{assumption}{Assumption}
\newtheorem{proposition}{Proposition}

\newcommand{\bea}{\begin{eqnarray}}
\newcommand{\eea}{\end{eqnarray}}

\newcommand{\beas}{\begin{eqnarray*}}
	\newcommand{\eeas}{\end{eqnarray*}}

\newcommand{\abs}[1]{\left\vert #1 \right\vert}
\newcommand{\norm}[1]{\left\Vert#1\right\Vert}
\renewcommand{\baselinestretch}{0.95}

\begin{document}

\title{Optimizing Basis Function Selection in Constructive Wavelet Neural Networks and Its Applications}
\author{Dunsheng~Huang, Dong~Shen, Lei~Lu, and Ying Tan
\thanks{This work was supported by the National Natural Science Foundation of China (62173333) and Beijing Natural Science Foundation (Z210002).}
\thanks{D. Huang and D. Shen are with the School of Mathematics and Research Center for Applied Mathematics, Renmin University of China, Beijing 100872, P.R. China (e-mail: dshuang@ruc.edu.cn, dshen@ieee.org).}
\thanks{L. Lu is with School of Life Course \& Population Sciences, King's College London, UK; Department of Engineering Science, University of Oxford, UK  (e-mail:
lei.lu@kcl.ac.uk).}
\thanks{Y. Tan is with the Department of Mechanical Engineering, The University of Melbourne, Parkville, VIC 3010, Australia (e-mail: yingt@unimelb.edu.au).}
}

\markboth{Manuscript for IEEE Transactions}{Short Title of Manuscript}

\maketitle

\begin{abstract}
Wavelet neural network (WNN), which learns an unknown nonlinear mapping from the data, has been widely used in signal processing, and time-series analysis. However, challenges in constructing accurate wavelet bases and high computational costs limit their application. This study introduces a constructive WNN that selects initial bases and trains functions by introducing new bases for predefined accuracy while reducing computational costs. For the first time, we analyze the frequency of unknown nonlinear functions and select appropriate initial wavelets based on their primary frequency components by estimating the energy of the spatial frequency component. This leads to a novel constructive framework consisting of a frequency estimator and a wavelet-basis increase mechanism to prioritize high-energy bases, significantly improving computational efficiency. The theoretical foundation defines the necessary time-frequency range for high-dimensional wavelets at a given accuracy. The framework's versatility is demonstrated through four examples: estimating unknown static mappings from offline data, combining two offline datasets, identifying time-varying mappings from time-series data, and capturing nonlinear dependencies in real time-series data. These examples showcase the framework's broad applicability and practicality. All the code will be released at https://github.com/dshuangdd/CWNN.
\end{abstract}

\begin{IEEEkeywords}
Constructive wavelet neural networks, reduced computational complexity,  initial wavelet frequency estimation, wavelet-basis increase mechanism.
\end{IEEEkeywords}

\section{Introduction}
\label{sec:introduction}

\IEEEPARstart{N}{eural} networks (NN), capable of learning complex data patterns, are widely used in applications \cite{zhou2023semi}. However, designing optimal architectures often requires a time-consuming trial-and-error process, even for experienced researchers \cite{shen2024autonet}. Constructive learning offers an efficient framework for incrementally building near-optimal architectures by starting with a small network and expanding it dynamically until satisfactory performance is achieved \cite{parekh2000constructive}. This approach has been applied to various NNs such as probabilistic NN \cite{Diamond1998}, cascade NN \cite{knorozova2024expressivity}, and extreme learning machines \cite{Incremental2015}.

Wavelet neural networks (WNNs) are specialized neural networks that use wavelet basis functions to learn unknown nonlinear mappings from data. WNNs leverage orthonormal wavelet functions for localization in both input and frequency domains, enabling hierarchical and multiresolution learning \cite{Alexandridis2013}. This makes them effective for tasks involving experimental data and diverse engineering applications \cite{9511133}. Multidimensional WNNs retain the ``universal approximation'' property of NNs and incorporate tailored wavelet bases for specific applications \cite{Zhang_1995,mohan2020wavelet}.

These properties of WNN make it ideal for constructive learning \cite{Ho}. WNNs' multiscale analysis captures both coarse and fine-grained patterns in data. Constructive learning leverages this by starting broadly and adding detail progressively \cite{24}. In WNNs, higher layers refine lower-layer outputs hierarchically. Algorithms can add layers or nodes to refine initial features, improving model precision and robustness \cite{parekh2000constructive}. Combining WNNs' multiresolution and hierarchy with constructive learning's incremental strategy creates efficient architectures. Starting simple and adding wavelet functions dynamically adjusts network complexity, ensuring robust performance with minimal computation overhead \cite{parekh2000constructive}. This approach accelerates NN design and enhances generalization \cite{mohan2020wavelet}.

Previous studies have leveraged the orthogonality of wavelet bases for constructive learning. For example, research in \cite{782410} developed structure-adaptive WNNs, selecting hidden layer nodes based on performance needs. Key areas include initialization structures \cite{17}, wavelet selection (e.g., residual-based methods \cite{yang2023wavecnns}, adaptive wavelet control \cite{24}, function approximation \cite{xu2011adaptive}), and parameter tuning/training \cite{He2002}. Training methods like Levenberg-Marquardt and quasi-Newton have been applied for new wavelet bases \cite{Ho}, while backpropagation has been used for network training \cite{Zhang2007}. These approaches are widely used in applications such as wind-induced vibration modeling \cite{xu2023physics} and electrical resistivity imaging inversion \cite{yu2023research}.
Recently, wavelets have been used in multiwavelet neural operators \cite{gupta2021} to solve PDEs like Burgers and Navier-Stokes equations via multi-scale sparse representations and their variants \cite{gupta2022, xiao2023coupled, Tapas2023, yuRixin2025}.

Despite their success, constructive WNNs face key challenges: (i) they require tuning many hyperparameters—e.g., neuron addition criteria, wavelet selection, and learning rates—which is time-consuming \cite{24}; (ii) they struggle with high-dimensional inputs due to the complexity of constructing wavelet bases \cite{Alexandridis2013}; and (iii) They under-utilize wavelets' time (spatial)-frequency localization—a key strength in signal processing, yet largely untapped for learning nonlinear mappings from data\cite{Zhang2009}.

This study presents a constructive WNN framework to estimate the frequency content of unknown nonlinear mappings. Using symmetric mother wavelets, it extracts spatial frequency components from known mappings and estimates dominant frequencies of unknown ones by evaluating energy in the wavelet subspace—guiding the selection of initial wavelet frequencies.

Corollary \ref{coro1} shows that many wavelet coefficients approach zero, enabling an iterative algorithm that systematically adds bases with the highest estimated energy. This leads to efficient WNN construction and refinement. The framework is validated through four cases: (1) learning with minimal bases to avoid over-parameterization \cite{Su_Lili_2019}; (2) combining offline datasets; (3) learning from time-series data; and (4) modeling nonlinear dependencies in real-world time-series applications.

The contributions of this study are summarized as follows.
\begin{enumerate}
\item This work introduces a novel approach to {\em analyzing} and {\em estimating} the frequency components of {\bf unknown} nonlinear functions and selecting an appropriate initial wavelet for approximation. By determining the main frequency components, the method can be applied to various types of nonlinear functions, significantly enhancing computational efficiency.

\item This work also provides a theoretical foundation for developing an iterative algorithm that systematically increases the wavelet bases using frequency analysis.
    
\item The proposed method has been validated across a range of scenarios for constructive learning, including both online and offline nonlinear mappings, as well as nonlinear dynamic systems. 
\end{enumerate}

The remainder of this paper is organized as follows. Section \ref{sec2} presents the problem formulation, WNN overview with revisions, and objectives. Section \ref{III-A} details the selection of initial wavelets, while Section \ref{III-B} describes the mechanism for increasing the number of wavelets to achieve desired accuracy. Section \ref{sec5} provides five numerical examples to validate the results. Section \ref{discussion} discusses the distinctions of the proposed framework. Finally, Section \ref{sec6} concludes the paper.

\section{Problem Formulation and Preliminaries}\label{sec2}

\subsection{Notations}
We denote by $\textbf{R}$ and $\textbf{R}^d$ the real space and a real d-dimensional space, respectively. The notation $\boldsymbol{x} \in \textbf{R}^d$ denotes a $d$-dimensional vector, i.e., $\boldsymbol{x}=[x_1,x_2,\ldots,x_d]^T$, where $(\cdot)^T$ denotes the transpose. For any vector $\boldsymbol{x} \in \textbf{R}^d$, $\abs{\boldsymbol{x}}$ represents the Euclidean norm, defined as $\abs{\boldsymbol{x}} \:= \sqrt{\boldsymbol{x}^T \boldsymbol{x}}$. For any vector $\boldsymbol{x} \in \textbf{R}^d$, $|\boldsymbol{x}|_s$ denotes the absolute value of $\boldsymbol{x}$, defined as $|\boldsymbol{x}|_s = [|x_1|,|x_2|,\dots,|x_d|]^T$. $\textbf{Z}$ and $\textbf{Z}^d$ denote the sets of integers and d-dimensional integers, respectively. For functions $f(\boldsymbol{x}),g(\boldsymbol{x}): \textbf{R}^d\rightarrow\textbf{R}$, $\left\langle{f,g}\right\rangle\triangleq \int_{\textbf{R}^d}f(\boldsymbol{x})g(\boldsymbol{x})d\boldsymbol{x}$ denotes the inner product. $C^d\left[0,\mathcal{T}'\right]$ denotes the $d$-order differentiable and continuous functions over the interval $\left[0,\mathcal{T}'\right]$. ${\partial y(t)}/{\partial{t}}$ represents the derivative of $y(t)$ with respect to $t$. For a $d$-dimensional vector $\boldsymbol{x}=[x_1,x_2,\ldots,x_d]^T\in\textbf{R}^d$, we use $\boldsymbol{x}\succeq0$ to denote $x_i>0$, $\forall 1\leq i\leq d$. Moreover, for two vectors $\boldsymbol{x},\boldsymbol{y}\in\textbf{R}^d$, we use $\boldsymbol{x}\succeq \boldsymbol{y}$ to denote $x_i\geq y_i$, $\forall 1\leq i\leq d$. Let $\boldsymbol{y} = \{y_1,y_2,...,y_n\}$ be a set with $n$ elements. For a given value $x\in\textbf{R}$, let
    $y^\ast = \mathop{\arg\min}_{\boldsymbol{y}} {\left\{ \left|y_i-x\right| \mid y_i \in \boldsymbol{y}, i=1,2,\dots,n\right\}}$. If there exist multiple candidates in  $\mathop{\arg\min}_{\boldsymbol{y}} \{ |y_i-x| \mid y_i \in \boldsymbol{y}, i=1,2,\dots,n\} $, we choose one randomly as $y^\ast$. Let $y^{\Delta} = \mathop{\arg\min}_{\boldsymbol{y} \backslash y^\ast} \{ |y_i-x| \mid y_i \in \boldsymbol{y}, i=1,2,\dots,n\}$. If there exist multiple candidates in $\mathop{\arg\min}_{\boldsymbol{y} \backslash y^\ast} \{ |y_i-x| \mid y_i \in \boldsymbol{y}, i=1,2,\dots,n\}$, we choose one randomly as $y^{\Delta}$, where ${\boldsymbol{y} \backslash y^\ast}$ represents removing element $y^\ast$ from $\boldsymbol{y}$, leading to  $\mathscr{U} (x,\boldsymbol{y}) = \{ y^\ast,y^{\Delta} \} $.

\subsection{System Model}

Consider the following nonlinear static mapping:
\begin{align}\label{syst_1}
y = f({\boldsymbol{x}}),
\end{align}
where ${\boldsymbol{x}}=\left[x_{1},x_{2},\dots,x_{d}\right]^T\in \textbf{R}^d$ denotes the input of the mapping, and $d$ represents the dimension of the input ${\boldsymbol{x}}$. The nonlinear mapping $f(\cdot): \textbf{R}^d\rightarrow \textbf{R}$ is locally Lipschitz continuous over $\boldsymbol{x}\in \textbf{R}^d$.

\begin{assumption}\label{assum1}
Suppose the nonlinear function $f(\cdot) \in L^2(\textbf{R}^d)$, meaning it is square-integrable over $\textbf{R}^d$.
\end{assumption}

Countable wavelet-basis functions form an orthonormal basis in $L^2(\textbf{R}^d)$. In this study, $f$ is restricted to the space $L^2(\textbf{R}^d)$, where it can be approximated by a finite number of wavelet bases to any desired precision \cite{17,16}. For $f\notin L^2(\textbf{R}^d)$, $\boldsymbol{x}$ is typically bounded within a finite interval due to the finite mapping. A compact set $D\subset \textbf{R}^d$ exists such that $f\in L^2(D)$ with $\boldsymbol{x}\in D$. Thus, Assumption \ref{assum1} is critical for the wavelet-network approximation framework.

The nonlinear function in \eqref{syst_1} is often transformed into products of unknown parameters and known basis functions. Existing methods typically rely on function approximation, with multiresolution analysis of orthogonal wavelet functions being particularly effective for nonlinear approximations \cite{16,21}. The orthogonality of wavelet bases allows continuous addition to gradually improve approximation accuracy until the desired level is reached. This property ensures optimal accuracy with a given number of basis functions. However, in high-dimensional cases, the curse of dimensionality leads to an excessive number of basis functions, complicating their effective use. Additionally, selecting an initial set of wavelets for unknown nonlinear mappings can be challenging, especially in control design applications for dynamic systems, where poor choices may result in unstable closed-loop behavior.

This study proposes a constructive WNN using basis selection to achieve predefined accuracy with fewer basis functions. It employs symmetric wavelet bases and spatial domain frequency analysis in $\textbf{R}^d$ to develop an algorithm for initial wavelet selection and a mechanism for adding new wavelets. 

\subsection{Wavelet approximation and frequency analysis}

This subsection begins with wavelet basis for functional approximation in $\textbf{R}^d$, followed by frequency analysis of basis functions from one-dimensional to multi-dimensional cases, establishing the foundation for this work.

\underline{Using wavelet basis functions to approximate $f(\cdot)$:}
Let ${\psi}_j(\boldsymbol{x}) \in \mathcal{S}$ (for $j = 1, 2, \dots$) be the basis of a normalized linear space $\mathcal{S}$ on $\textbf{R}^d$. For any $f \in \mathcal{S}$ satisfying Assumption \ref{assum1}, we have $f(\boldsymbol{x}) = \sum_{j=1}^{N_\mathcal{S}} {\theta}_j {\psi}_j(\boldsymbol{x})$, where $N_\mathcal{S}$ can be infinite and ${\theta}_j \in \textbf{R}$ are coefficients. If $\mathcal{S} = L^2(\textbf{R}^d)$, ${\psi}_j(\boldsymbol{x})$ can represent local Lipschitz continuous (LLC) wavelet-basis functions. In practice, $f(\cdot)$ is approximated using a finite number of wavelet bases to achieve desired accuracy, ensuring the approximation error norm is below a specified threshold \cite{28}.
The approximation is given as follows:
\bea
{\small {\mbox {$f(\boldsymbol{x})=\sum_{j=1}^N{{\theta }_j{\psi }_j(\boldsymbol{x}) }+{e}_N(\boldsymbol{x})$}}},
\eea
where $N$ represents the number of bases in the approximation and ${e}_N(\boldsymbol{x}) $ is the approximation error, satisfying
\beas
{\small{\mbox{$\lim_{{N \to \infty}} \norm{e_N}^2 = \lim_{{N \to \infty}} \int_{\boldsymbol{x}\in \textbf{R}^d} {|e_N(\boldsymbol{x})|^2 d\boldsymbol{x}} = 0$}}}.
\eeas
Here $\norm{\cdot}^2$ represents the $L^2$ norm. As $\psi_j(\boldsymbol{x})$ is an orthogonal wavelet basis, $\theta_j$ is unique \cite{16}.

Since the input $\boldsymbol{x}$ in mapping \eqref{syst_1} is $d$-dimensional, we construct $d$-dimensional wavelet functions using the \textbf{Single-Scaling Wavelet Frame} method from \cite{26}, discretized as follows:
\bea
\label{eq2}
\small{\mbox{$\mathcal{W}\triangleq\left\{\psi_{mn}(\boldsymbol{x})=2^{\frac{dm}{2}}\psi(2^{m}\boldsymbol{x}-n)|m \in \textbf{Z}, n \in \textbf{Z}^d,\boldsymbol{x}\in \textbf{R}^d\right\}$}},
\eea
where $\textbf{Z}^d$ denotes the set of $d$-dimensional integers. The wavelet functions $\{\psi_{mn}(\boldsymbol{x})\}$ are generated by scaling and translating a single mother wavelet $\psi(\cdot)$, with $m$ and $n$ controlling the scaling and translation, respectively. Thus, these wavelet functions are derived from a single mother wavelet. Specifically, the translation center and frequency center for these wavelets are $2^{-m}n$ and $2^m$, respectively \cite{16}.

This study focuses on a special form of the mother wavelet, $\psi(\boldsymbol{x})$, with a well-defined Fourier transform $\widehat{\psi}(\boldsymbol{\omega})$ in $\textbf{R}^d$. The terms $\widehat{\psi}(\boldsymbol{\omega})$ and $\psi(\boldsymbol{x})$ form a Fourier pair. We revisit the process of generating $\psi(\boldsymbol{x})$ from one-dimensional to $d$-dimensional cases as described in \cite{26}.

The mother wavelet $\psi(\boldsymbol{x})$ is obtained by the inverse Fourier transform of $\widehat{\psi}(\boldsymbol{\omega})$. To determine $\widehat{\psi}(\boldsymbol{\omega})$, we first select a function $\phi(x)$ and its Fourier transform $\widehat{\phi}(\omega)$ for scalars $x$ and $\omega$ that satisfy the following three conditions:
\bea
\label{neweq1}
   &\ & \phi(x)=\phi(-x) \\
\label{eq4}
    &\ &{\small \mbox{$\inf_{|\omega|\in[1,2]}\sum_{m\in \textbf{Z}}\left|\widehat{\phi}\left(2^{-m}\omega\right)\right|^2\ > 0$}}\\
\label{eq5}
    &\ & {\small \mbox{$\left|\widehat{\phi}(\omega)\right| \leq C|\omega|^{\alpha}\left(1+|\omega|^2\right)^{-\frac{\gamma}{2}}$}},
\eea
where $C > 0$, $\alpha > 0$, and $\gamma >\alpha+d$ are constants. 
Many functions satisfy these conditions, including the Sinc function and one-dimensional wavelet functions with exponential decay.

Once $\widehat{\phi}\left(\omega\right)$ and ${\phi}\left(x\right)$ are obtained, the $d$-dimensional  $\widehat{\psi}(\boldsymbol{\omega})$ can be obtained by using the following relation \cite{26}:
\begin{align}
\label{eq3}
    \widehat{\psi}(\boldsymbol{\omega})=\widehat{\phi}\left(\abs{\boldsymbol{\omega}}\right),
\end{align}
where $\boldsymbol{\omega}\in \textbf{R}^d$. The mother wavelet $\psi(\boldsymbol{x})$ can be obtained using an inverse Fourier transform. The following example demonstrates how to generate $\psi(\boldsymbol{x})$ \cite{26}.

\emph{Example:} The one-dimensional Mexican Hat function ${\phi}(x) = (1-x^2)e^{-x^2/2}$ and its Fourier transform $\widehat{\phi}(\omega) = \omega^2 e^{-\omega^2/2}$ satisfy conditions \eqref{neweq1}, \eqref{eq4}, and \eqref{eq5}. Using \eqref{eq3}, we derive the $d$-dimensional function $\widehat{\psi}(\boldsymbol{\omega}) = |\boldsymbol{\omega}|^2 e^{-|\boldsymbol{\omega}|^2/2}$. Applying the inverse Fourier transform yields the $d$-dimensional mother wavelet $(d - |\boldsymbol{x}|^2)e^{-|\boldsymbol{x}|^2/2}$, which can be used in \eqref{eq2} to construct wavelet functions.

When using WNN, as dimensionality $d$ increases, the number of wavelet basis functions also rises significantly, creating challenges for engineering applications. This study aims to develop a practical method to reduce the number of wavelet basis functions while maintaining the required approximation accuracy for the unknown nonlinear function $f(\cdot)$ in \eqref{syst_1}, which satisfies Assumption \ref{assum1}, using input and output measurements.

{\underline {Objective of this work:}} For a given collected  dataset $\{\boldsymbol{x}_k, y_k\}_{1\leq k\leq N}$, we want to approximate an unknown mapping $f(\cdot)$ by constructing a WNN systematically. It   starts with a carefully design initial WNN and increasing the number of wavelets until the desired accuracy is achieved. Specifically, given $\varepsilon > 0$, the goal is to design an algorithm that increases the number of wavelets from the carefully designed initial WNN to ${N_\theta}$ wavelets, ensuring the difference between predicted and actual outputs is less than $\varepsilon$ for all data points:
\bea
{\small \mbox{$\displaystyle \sum_{k=1}^{N}\displaystyle \left( \sum_{j=1}^{N_\theta}{\theta }_j{\psi }_j(\boldsymbol{x}_k)-y_k\right)^2\leq \varepsilon$}}.
\eea
The method has two key components: an initial frequency estimator that localizes the primary energy of $f(\cdot)$ to reduce computation, and a basis expansion mechanism that adds high-energy wavelets based on estimated coefficients.

\section{Estimation of the Initial Frequency}\label{III-A}

For an unknown nonlinear function $f$, we first determine the initial wavelet subspace and its corresponding frequency for approximation. Many studies rely on empirical experience to select a wavelet subspace and use all its bases to approximate $f$ \cite{25}. However, this approach can waste computational resources or slow convergence. For example, when approximating complex nonlinear functions with high-frequency components, selecting too low an initial frequency slows convergence, while choosing an excessively high frequency wastes resources.

This work provides a theoretical basis for selecting an appropriate initial frequency. According to \cite{16}, the function $f(\cdot)$ can be projected onto wavelet subspaces and approximated using their bases. The total energy is the sum of the energies in each subspace. By identifying subspaces from the measured data that capture most of the energy of $f(\cdot)$, we can effectively approximate the unknown nonlinear mapping using the corresponding bases. The frequencies associated with these predominant subspaces, termed main frequencies, are selected as the initial wavelet frequency.

\begin{figure}[!t]
\centering
\includegraphics[width=0.3\textwidth]{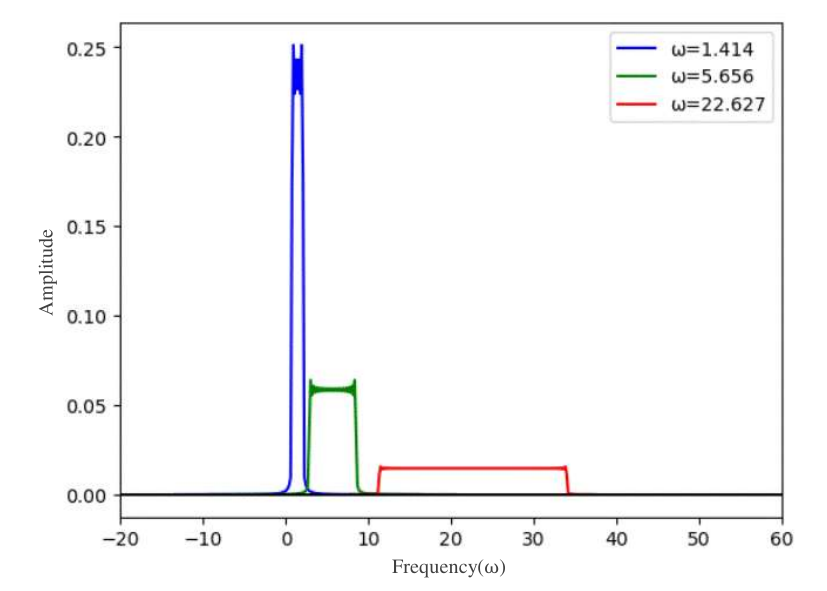}\vspace{-2mm}
\caption{Frequency spectrum of three Sinc wavelets with different central frequencies $\omega$}
\label{fig2}
\end{figure}

Let $E_m$ represent the energy of the nonlinear function $f(\cdot)$ projected onto the wavelet subspace ${W}_m$. First, we show how to compute $E_m$ for a known nonlinear mapping. Then, we explain how to estimate $E_m$ using the collected data $\left\{\boldsymbol{x}_k, y_k\right\}_{1\leq k\leq N}$.

\subsection{Computing $E_m$ for a known $f(\cdot)$}

The following proposition provides a method for calculating the energy $E_m$ of wavelet subspace ${W}_m$.
\begin{proposition}\label{proposition1}
Assuming that the projection of $f(\boldsymbol{x})$ on the wavelet space ${W}_m$ is ${f}_m(\boldsymbol{x})$, the energy $E_m$ of ${f}_m(\boldsymbol{x})$ is calculated as follows:
\begin{align}\label{energy_E_m}
E_m=\sum_n{{C^2_{mn}\|{\psi}_{mn}(\boldsymbol{x})\|}^2},
\end{align}
where $C_{mn}={\frac{\int_{\textbf{R}^d}{f(\boldsymbol{x})\cdot{\psi }_{mn}(\boldsymbol{x})\ d{\boldsymbol{x}}}}{\int_{\textbf{R}^d}{{\psi }_{mn}(\boldsymbol{x})\cdot{\psi }_{mn}(\boldsymbol{x})\ d{\boldsymbol{x}}}}}$ denotes the coefficients. 
\end{proposition}
{\em Proof}: See the Supplementary Material. \hfill $\Box$

When identifying primary energy subspaces, we estimate the wavelet subspace energy across all frequencies using measurement data, especially when the nonlinear mapping is unknown. If the energy $E_m$ of subspace $W_m$ has a unimodal distribution with respect to $m$, the main energy subspaces are near the peak, allowing estimation from the data. If $E_m$ has a multimodal distribution and the number of peaks is unknown, identifying the specific peak for the main energy subspaces becomes difficult. To simplify analysis, we make the following assumption.
\begin{assumption}\label{assum2}
$E_m$ is unimodal concerning resolution $m$.
\end{assumption}
\begin{remark} The frequency bandwidth of a wavelet function increases with frequency \cite{16}. Fig.\,\ref{fig2} shows the frequency spectrum from the Fourier transform of three Sinc wavelets with varying central frequencies $\omega$. Higher central frequencies correspond to wider bandwidths. As a result, more components of $f$ fall into higher-frequency subspaces, which typically contain more energy. Initially, the energy $E_m$ increases with $m$, but due to the limited total energy of $f$, it eventually decreases if $E_{m+1} < E_m$, forming a unimodal distribution. Thus, Assumption 2 is not restrictive, as demonstrated in simulations.\hfill $\circ$
\end{remark}

Assumption \ref{assum2} states that energy $E_m$ has a single peak with respect to resolution $m$. To identify the main energy subspaces more efficiently, we introduce an energy evaluation index using the Exponential Moving Average (EMA) method to recursively determine $E_m$.

Letting $E_m$ be the original energy computed by \eqref{energy_E_m}, the EMA-transformed $E_m^\circ$ is obtained as follows:
\bea
E_m^\circ=\frac{\alpha E_{m-1}^\circ+\left(1-\alpha \right)E_m}{1-{\alpha }^m}, m=2,3,\ldots \label{EMA_definition}
\eea
where ${\alpha \in [0,1)}$ and $E_1^\circ\triangleq E_1$. 
The parameter $\alpha$ reflects the importance of the previous space $E_{m-1}$. A larger $\alpha$ indicates that $E_m^\circ$ contains more energy information at resolutions smaller than $m$.

The selection of the parameter is case-dependent. A smaller accuracy parameter $\varepsilon$ means fewer wavelets are truncated, resulting in a smaller recursive average $E_m^\circ$. This parameter is computed using the following empirical rule:
\bea
\alpha =\frac{2{\mathrm{arctan} \left(-{{\lg} (\mathrm{\varepsilon}) }\right)}}{\pi},
\label{definition_alpha}
\eea
where $\lg(\cdot)$ is the base-10 logarithm and $\varepsilon$ is the approximation accuracy, $\lg(\varepsilon)$ increases resolution for small $\varepsilon$, while $2{\arctan}(\cdot)/\pi$ maps $(0,\infty)$ to $(0,1)$ to match $\alpha \in (0,1)$.
Consequently, it has
\bea
\label{eq11}
{\small {\mbox {$E_m^\circ=\dfrac{{\mathrm{arctan}\left(-{\mathrm{lg} \left(\mathrm{\varepsilon}\right)}\right)}E_{m-1}^\circ+\left(\frac{\pi }{2}-{\mathrm{arctan} \left(-{\mathrm{lg} \left(\mathrm{\varepsilon }\right)}\right)}\right)E_m}{\frac{\pi}{2}\left(1-{\left(\frac{2{\mathrm{arctan} \left(-{\mathrm{lg} \left(\mathrm{\varepsilon }\right) }\right) }}{\pi }\right)}^m\right)}$}}}.
\eea
If $E_{m+1} \leq E_m^\circ$, the energy in space $W_{m+1}$ is sufficiently small to terminate the EMA calculation. Here, $E_m^\circ$ approximates the average of the latest $\lceil{1/(1-\alpha)}\rceil$ data points of $E_m$, i.e., $\{E_j\}_{j={\lceil{1/(1-\alpha)}\rceil},\dots,m}$, where $\lceil{\cdot}\rceil$ is the ceiling function. This recursive process ensures that $W_{m+1}$ likely contains less energy, showing that $E_m^\circ$ is more suitable than $E_m$.

\subsection{Estimating $E_m$ for an unknown $f(\cdot)$}

Given the unknown function $f(\cdot)$, computing $C_{mn}$ in \eqref{syst_1} or $E_m$ in \eqref{EMA_definition} is not feasible. Thus, we estimate the energy ${\widehat{E}}_m$ using the measured data $f(\cdot)$.

The energy of the function $f(\cdot)$ in each wavelet subspace can be estimated using wavelet bases. Denote the estimated coefficients as $\widehat{C}_{mn}$ and the energy as $\widehat{E}_m$. Then, we obtain
\bea
\label{eq10}
\widehat{E}_m=\displaystyle \sum_n{{\widehat{C}^2_{mn}\|{\psi}_{mn}(\boldsymbol{x})\|}^2}.
\eea
This leads to the estimated energy ${\widehat{E}}_m$ as follows:
\bea
\overline{E}_m=\frac{\alpha \overline{E}_{m-1}+\left(1-\alpha \right)\widehat{E}_m}{1-{\alpha }^m},
\eea
where $\alpha$ is defined in (\ref{definition_alpha}).
In this work, Algorithm \ref{alg1} is employed for this estimation.

In \emph{Step 1}, select $N_1$ bases in space $W_1$, estimate its energy $\widehat{E}_{1}$, and compute the average $\overline{E}_{1}$. In \emph{Step 2}, select corresponding bases in $W_2$, and calculate their average $\overline{E}_{2}$ using $\overline{E}_{1}$ and $\hat{E}_{2}$. Finally, in \emph{Step 3}, decide whether to terminate or continue iterating \emph{Step 2} with higher-frequency bases.

\begin{algorithm}[t]
\label{alg1}
\SetAlgoLined
\caption{Estimating the initial wavelet frequency}
\KwIn{The translation centers $\{{\boldsymbol{K}}_{1,1},{\boldsymbol{K}}_{1,2},\dots,{\boldsymbol{K}}_{{1,N_1}}\}$ of the wavelet bases in space $W_{1}$, the hyperparameter $\kappa$, the learning rate $\iota_r$, the input data $\{\boldsymbol{x}_i\}$ and actual output data $\{y{(\boldsymbol{x}_i)}\} (i=1,2,\dots,N_{sa})$.}
\KwOut{Initial wavelet frequency $2^{m_{init}}$.}
{Select the translation centers {$\{{\boldsymbol{K}}_{1,1},{\boldsymbol{K}}_{1,2},\dots,{\boldsymbol{K}}_{1,{N}_{J}}\}$} in $W_{1}$; calculate the coefficients $\widehat{C}_{1n}$ by \eqref{neweq2} and then compute ${\widehat{E}}_{1}$ and ${\overline{E}}_{1}$\;}
{Select the translation centers {$\{{\boldsymbol{K}}_{2,1},{\boldsymbol{K}}_{2,2},\dots,{\boldsymbol{K}}_{2,2^d*{N}_{J}}\}$} in $W_{2}$; calculate the coefficients $\widehat{C}_{2n}$ by \eqref{neweq3} and then compute ${\widehat{E}}_{2}$ and ${\overline{E}}_{2}$\;}
{$m\leftarrow 2$\;}
\While{${\overline{E}}_{m}\le {\widehat{E}}_{m+1}$}{
$m\leftarrow m+1$\;
{\rm{Select the translation centers} {$\{{\boldsymbol{K}}_{2,1},{\boldsymbol{K}}_{2,2},\dots,{\boldsymbol{K}}_{2,2^{dm}*{N}_{J}}\}$} in $W_{m+1}$; calculate the coefficients $\widehat{C}_{(m+1)n}$ and then compute $\widehat{E}_{m+1}$ and $\overline{E}_{m+1}$\;}
}
\rm{Stop at} $m_{init}$;
\end{algorithm}

\begin{figure*}[t]
\centerline{\includegraphics[width = 0.6\textwidth]{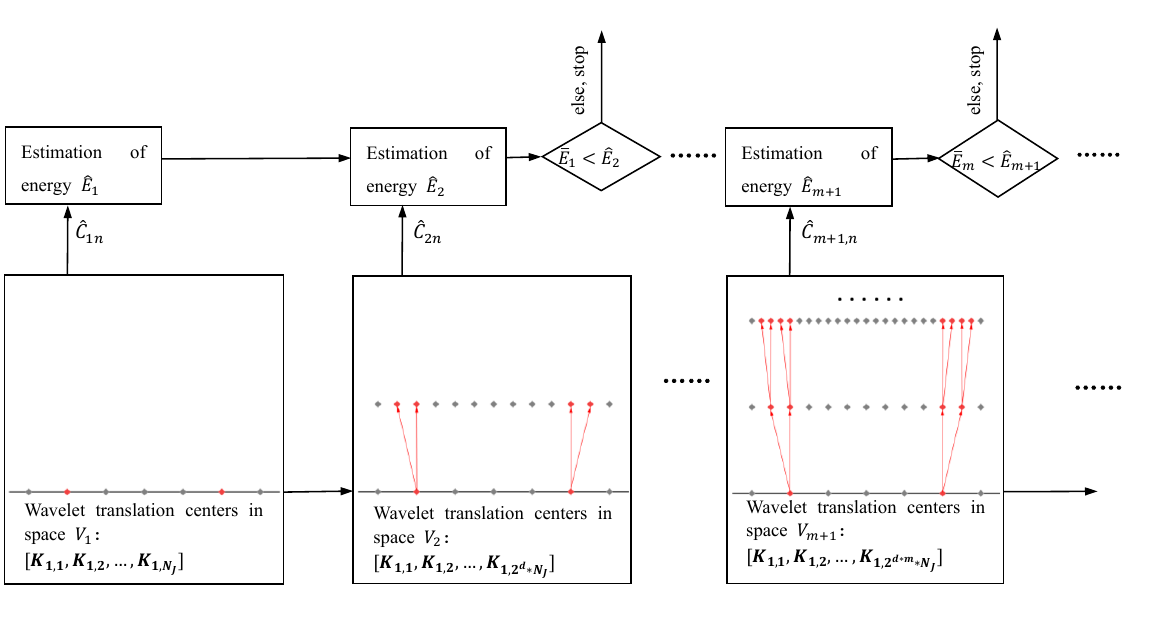}}
\caption{Flowchart for selecting the initial wavelet frequencies (red dots indicate the wavelet bases chosen in the relative space)}
\label{fig3}
\end{figure*}

\emph{Step 1:} The translation centers of the wavelet bases in space $W_{1}$ are denoted as $\{{\boldsymbol{K}}_{1,1},{\boldsymbol{K}}_{1,2},\dots,{\boldsymbol{K}}_{{1,N_1}}\}$, where ${\boldsymbol{K}}_{1,1},{\boldsymbol{K}}_{1,2},\dots,{\boldsymbol{K}}_{{1,N_1}}\in \textbf{R}^d$ and $N_1$ is the number of bases. $\{{\boldsymbol{K}}_{1,1},{\boldsymbol{K}}_{1,2},\dots,{\boldsymbol{K}}_{{1,N_J}}\}$ are elements selected from $\{{\boldsymbol{K}}_{1,1},{\boldsymbol{K}}_{1,2},\dots,{\boldsymbol{K}}_{{1,N_1}}\}$ and $N_J=\kappa \cdot N_1$, where the hyper-parameter $\kappa \in (0,1)$ is usually small to save computational resources. The corresponding wavelet bases are denoted as ${\boldsymbol{\psi }}_{1}=[{\psi }_{1,1},{\psi }_{1,2},\dots,{\psi }_{1,N_J}]^T$.

There are various methods to determine $\widehat{C}_{1n}$. In this study, a three-layer NN (input, hidden, and output layers) is used to estimate $\widehat{C}_{1n}$ from training data consisting of input $\{\boldsymbol{x}_i\}$ and actual output $\{y(\boldsymbol{x}_i)\}$ ($i=1,2,\dots,N_{sa}$), where $N_{sa}$ is the number of data points. The initial wavelet coefficients are set to $0$, i.e., ${{\widehat{C }}^o_{1n}} = 0$ for all $n=1,2,\dots, N_J$. After importing the training data into the network and running it once, the updated basis coefficients ${{\widehat{C }}_{1n}}$ ($n=1,2,\dots, N_J$) are obtained as follows:
\bea
\label{neweq2}
{{\widehat{C }}_{1n}} = {{\widehat{C }}^o_{1n}} + \iota_r \frac{2}{N_{sa}}\sum^{N_{sa}}_{i=1}(y(\boldsymbol{x}_i)-\widehat{y}(\boldsymbol{x}_i))
\psi_{1n}(\boldsymbol{x}_i),
\eea
where $\widehat{y}(\boldsymbol{x}_i)$ is the output corresponding to the $i^{th}$ input data, and $\iota_r$ is the learning rate. 

Applying \eqref{eq10} with the coefficients ${{\widehat{C }}_{1n}}(n=1,2,\dots, N_J)$, the energy estimate ${\widehat{E}}_{1}$ of $f$ in subspace $W_{1}$ is obtained as ${\widehat{E}}_{1}=\sum_{n=1}^{N_J}{({\widehat{C }}^{\ 2 }_{1n}){\|{\psi }_{1n}\|}^2}$.
We then set ${\overline{E}}_{1}={\widehat{E}}_{1}$.

\emph{Step 2:} At resolution $m=2$, select the $2^d$ bases from the wavelet subspace $W_{2}$ with translation centers closest to $\boldsymbol{K}\in\{{\boldsymbol{K}}_{1,1},{\boldsymbol{K}}_{1,2},\dots,{\boldsymbol{K}}_{{1,N}_{J}}\}$. After traversal, $2^{d}\cdot {N}_J$ bases are obtained with translation centers labeled as $\{{\boldsymbol{K}}_{2,1},\dots,{\boldsymbol{K}}_{2,2^d\cdot {N}_{J}}\}$. The corresponding wavelets are ${\boldsymbol{\psi }}_{2}=[{\psi }_{2,1},\dots,{\psi }_{2,2^d\cdot {N}_{J}}]^T$. Fig.\,\ref{fig3} illustrates this process for $d=1$. For the newly selected bases in $W_{2}$, we set all initial wavelet coefficients to $0$: ${{\widehat{C }}^o_{2n}} = 0, \forall n=1,\dots, 2^d\cdot {N}_{J}$.

Similar to \emph{Step 1}, the corresponding coefficients ${{\widehat{C}}_{2n}}(n=1,2,\dots, 2^d\cdot {N}_{J})$ are calculated as follows:
\bea
\label{neweq3}
{{\widehat{C }}_{2n}} = {{\widehat{C }}^o_{2n}} +\iota_r \frac{2}{N_{sa}}\sum^{N_{sa}}_{i=1}(y(\boldsymbol{x}_i)-\widehat{y}(\boldsymbol{x}_i))
\psi_{2n}(\boldsymbol{x}_i).
\eea
Then, ${\widehat{E}}_{2}$ is determined by \eqref{eq10}, and ${\overline{E}}_{2}$ is computed by \eqref{eq11}. If ${\overline{E}}_{1} \ge {\widehat{E}}_{2}$, the process terminates; otherwise, the algorithm proceeds to \emph{Step 2}.

\emph{Step 3:} Set $m=m+1$ and repeat \emph{Step 2} until ${\overline{E}}_m\ge {\widehat{E}}_{m+1}$. Select $2^{(m-1)}$ as the initial wavelet frequency $2^{m_{init}}$.

The method for determining the initial wavelet frequency is shown in Fig.\,\ref{fig3}. In \cite[Th.\,5]{28}, it is proven that wavelet transforms measure local function regularity. High regularity implies function smoothness, requiring high-frequency wavelets for accurate approximation. Thus, new bases in $W_{m}$ with translation centers near those in $W_{m-1}$ can be chosen to find higher-frequency bases resembling $f(\cdot)$. This insight guides \emph{Step 2} in Algorithm \ref{alg1}.

\begin{remark}
It is noted that during the initialization stage, the primary role is to determine the initial structure of the WNN. It is observed that increasing $m$, the number of layers, does not significantly improve the approximation energy. Moreover, in \emph{Step 2} of Algorithm~\ref{alg1}, the coefficients are updated only once after adding new wavelets to save computational resources. For offline approximation of $f(\cdot)$, coefficients can be updated multiple times to achieve better accuracy. In contrast, for online approximation, coefficients are updated only once due to time constraints.
\end{remark}

\begin{remark}
The parameter $\kappa$ in Step 1 of Algorithm 1 controls the number of translation centers in the subspace of $W_1$, thereby determining the quantity of selected bases in this subspace. The value of $\kappa$ is configured based on the affordable computational resources in practical applications: a larger $\kappa$ induces higher computational costs, yet yields a more accurate initial wavelet frequency. A reasonable recommended range for $\kappa$ is $\left(\frac{1}{3},\frac{2}{3}\right)$. In practice, values within this range have been observed to work well for a wide variety of functions, ensuring that the initial wavelet selection is neither too sparse (missing critical frequency components) nor too dense (wasting computational resources). 
\end{remark}

\section{Mechanism for Increasing Wavelet Bases}\label{III-B}

The number of wavelet bases increases rapidly from $W_{m}$ to $W_{m+1}$, even with frequency techniques. For example, increasing resolution from $m=4$ to $m=5$ adds 4,225 new bases in the simulation of \cite{25}, roughly tripling the original count. To reduce computational complexity, a mechanism is needed to add bases efficiently while maintaining approximation accuracy. Selecting bases with dominant energy offers key guidance. To systematically increase the number of wavelets, this subsection first examines the one-dimensional case and then extends to the high-dimensional case.

\subsection{One-dimensional function}

In this subsection, we revisit the result in \cite{16} with necessary definitions and notations.  
Suppose 
\bea
\left({\mathcal{Q}}_T f\right)(x)&=&{\chi }_{\left[-T,T\right]}(x)f({x}), \label{definition_Q_T}\\({\mathcal{P}}_{{\mathit{\Omega}}}{\widehat{f})\left({\omega} \right)}&=&{\chi }_{\left[-\mathit{\Omega},\mathit{\Omega}\right]}({\omega} )\widehat{f}\left({\omega} \right),\label{definition_P_Omega}
\eea
where $\widehat{f}(\omega)$ is the Fourier transform of $f(x)$, ${x} \in [-T,T]$, and ${\omega} \in [-\Omega,\Omega]$, with $x, \omega \in \mathbf{R}$ and $T, \Omega > 0$. For an interval $I$, the function ${\chi }_I$ is defined as 1 if $x \in I$ and 0 otherwise.

Although the nonlinear function $f(\cdot)$ may lack compact support in both frequency and time domains \cite{16}, we can identify compact sets in these domains where the main function energy is concentrated. This is referred to as the ``concentration'' of function energy. Specifically, for a bounded time domain $[-T, T]$ and frequency domain $[{\mathit{\Omega}}_0,{\mathit{\Omega}}_1]\cup[-{\mathit{\Omega}}_1,{-\mathit{\Omega}}_0]$, the energy ``concentration'' is defined within this domain:
\begin{align}
    &\|f\|\gg \|\left(\boldsymbol{1}-{\mathcal{P}}_{{\mathrm{\Omega }}_1}+{\mathcal{P}}_{{\mathrm{\Omega }}_0}\right)\widehat{f}\|,
    \nonumber\\
    &\|f\|\gg \|(\boldsymbol{1}-{\mathcal{Q}}_T)f\|,\nonumber
\end{align}
where ${\mathit{\Omega}}_0,{\mathit{\Omega}}_1 \in \textbf{R}$ are constants, ${\mathcal{Q}}_T f$ is defined in (\ref{definition_Q_T}), and ${\mathcal{P}}_{{\mathrm{\Omega }}_1}$ is defined in (\ref{definition_P_Omega}).  It is essential to approximate the nonlinear function $f$ over a finite range \cite{10263779,182697}.

In practice, the range of wavelet translation centers in the approximation is often set larger than $[-T, T]$ to achieve higher accuracy \cite{29}. If many wavelet coefficients are close to $0$, the number of wavelet bases can be reduced by omitting those with negligible coefficients.

\begin{figure}[t]
\centering
\subfigure[]{\label{fig4a}
\includegraphics[width=0.20\textwidth]{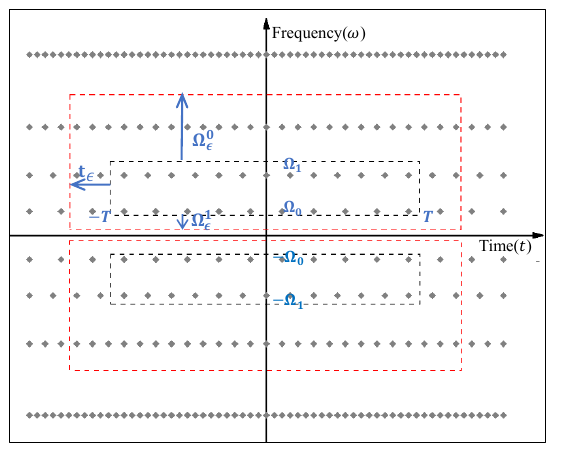}}
\subfigure[]{\label{fig4b}
\includegraphics[width=0.19\textwidth]{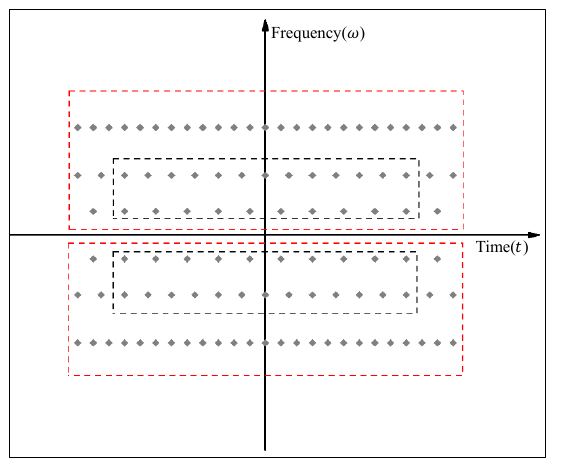}}
\caption{(a) Positions of wavelets $\{{\psi }_{mn}(x)\}$ in the time–frequency domain. The horizontal and vertical axes represent the wavelet translation and wavelet frequency, respectively. The black dashed box represents $B\left(T,{\mathrm{\Omega }}_0,{\mathrm{\Omega }}_1\right)$ and the red dashed box represents the enlarged $B_{\varepsilon }\left(T,{\mathrm{\Omega }}_0,{\mathrm{\Omega }}_1\right)$. (b) represents the domain excluding bases outside $B_{\varepsilon }\left(T,{\mathrm{\Omega }}_0,{\mathrm{\Omega }}_1\right)$.}
\label{fig4}
\end{figure}

To explain the relationship between the wavelet basis and $f$, we plot the time-frequency centers of the bases in Fig.\,\ref{fig4a}. Gray dots represent the centers of the wavelets ${\psi }_{mn}(x)$. The energy of $f$ is concentrated in the black dashed box $B(T, {\Omega}_0, {\Omega}_1) = [-T,T] \times ([{\Omega}_0, {\Omega}_1] \cup [-{\Omega}_1, -{\Omega}_0])$, where ${\Omega}_0 \leq 2^m \leq {\Omega}_1$ and $|n| \leq 2^m T$. The wavelet coefficient $C_{mn}$ decreases as the distance between the wavelet center and $B(T, {\Omega}_0, {\Omega}_1)$ increases. To achieve approximation accuracy $\varepsilon$, $B(T, {\Omega}_0, {\Omega}_1)$ must be expanded by adding wavelets ${\psi }_{mn}(x)$ from an extended time-frequency range.
The extended box $B_{\varepsilon }\left(T,{\Omega }_0,{\Omega }_1\right)=[-(T+t_{\varepsilon }),(T+t_{\varepsilon })]\times $([$({\Omega }_0-{\Omega}^0_{\varepsilon })$, (${\Omega }_1+{\Omega }^1_{\varepsilon })$]$\cup$[$-{(\Omega }_1+{\Omega }^1_{\varepsilon })$, ${-(\Omega }_0-{\Omega }^0_{\varepsilon })$]) (red box in Fig.\,\ref{fig4a}) satisfies
\bea
\label{eq14}
\left\|f-\sum_{\left(m,n\right)\in B_{\varepsilon }\left(T,{\Omega }_0,{\Omega }_1\right)}{\left\langle{\psi }_{mn},f\right\rangle{\psi }_{mn}}\right\|\boldsymbol{\le }\varepsilon,
\eea
where $t_{\varepsilon }, {\Omega}^0_{\varepsilon },{\Omega}^1_{\varepsilon }$
are positive constants related to $\varepsilon$.
The procedure for selecting these values can be found in \cite{16}.

Here, ${\mathrm{\Omega }}_0-{\mathrm{\Omega }}^0_{\varepsilon }\le 2^{m}\le {\mathrm{\Omega }}_1+{\mathrm{\Omega }}^1_{\varepsilon }$ and $|n|\le 2^{m}T+t_{\varepsilon }$. For a given $\varepsilon$, wavelet-basis coefficients outside $B_{\varepsilon }(T,{\mathrm{\Omega }}_0,{\mathrm{\Omega }}_1)$ approximate $0$, so these wavelets can be omitted without violating $\varepsilon$-accuracy. Thus, the bases for approximating $f$ are restricted to those within this region (see Fig.\,\ref{fig4b}).

\subsection{Multi-dimensional function}

We now generalize \eqref{eq14} to higher dimensions by extending the functions ${\mathcal{Q}}_T$ and ${\mathcal{P}}_{{\mathrm{\Omega}}}$ as follows:
\bea
\left({\mathcal{Q}}_{\boldsymbol{T}} f\right)(\boldsymbol{x})&=&{\chi }_{\left[-\boldsymbol{T},\boldsymbol{T}\right]}
(\boldsymbol{x})f({\boldsymbol{x}})\label{definition_Q_high_dimensional}\\
({\mathcal{P}}_{{\boldsymbol{\Omega}}}{\widehat{f})\left(\boldsymbol{{\omega}} \right)}&=&{\chi }_{\left[-\boldsymbol{{\Omega}},\boldsymbol{{\Omega}}\right]}({\boldsymbol{\omega}} )\widehat{f}\left({\boldsymbol{\omega}} \right)\label{definition_P_high_dimension}
\eea
where $\boldsymbol{x},\boldsymbol{\omega},\boldsymbol{T},\boldsymbol{ \Omega} \in \mathit{R}^d$, $\boldsymbol{T}\succeq0$, $\boldsymbol{\Omega}\succeq0$. The function $\chi_{[-\boldsymbol{T}, \boldsymbol{T}]}(\boldsymbol{x})$ is defined as $1$ if $\boldsymbol{x}\in [-\boldsymbol{T}, \boldsymbol{T}]$ and $0$ otherwise,
where $\boldsymbol{x}\in [-\boldsymbol{T}, \boldsymbol{T}]$ denotes $-\boldsymbol{T}\preceq \boldsymbol{x} \preceq \boldsymbol{T}$. The definition of $\chi_{[-\boldsymbol{\Omega}, \boldsymbol{\Omega}]}(\boldsymbol{\omega})$ is similar to ${\chi }_{[-\boldsymbol{T}, \boldsymbol{T}]}(\cdot)$.

Assumption \ref{assum3} is presented to facilitate the extension from the one-dimensional case \eqref{eq14} to higher dimensions.
\begin{assumption}\label{assum3}
    Let $\widehat{f}(\boldsymbol{\omega})$ be the Fourier transform of $f(\boldsymbol{x})$, where $\boldsymbol{\omega} \in \textbf{R}^d$ is the frequency. Assume that the energy of  $f(\boldsymbol{x})$ in \eqref{syst_1} is centralized in the time domain $[-\boldsymbol{T},\boldsymbol{T}]$ and frequency domain $[{\boldsymbol{\Omega}}_0,{\boldsymbol{\Omega}}_1]\cup[-{\boldsymbol{\Omega}}_1,{-\boldsymbol{\Omega}}_0]$; i.e., $\|f\|\approx\|({\mathcal{P}}_{{\boldsymbol{\Omega}}_1}-{\mathcal{P}}_{{\boldsymbol{\Omega}}_0} )\widehat{f}\|$, $\|f\|\approx\|{\mathcal{Q}}_{\boldsymbol{T}} f\|$, where $\boldsymbol{T}, \boldsymbol{\Omega}_0,\boldsymbol{\Omega}_1 \in \textbf{R}^d$.
\end{assumption}

This assumption concentrates the energy of $f(\boldsymbol{x})$ in the time domain $[-\boldsymbol{T},\boldsymbol{T}]$ and the frequency domain $[{\boldsymbol{\Omega}}_0,{\boldsymbol{\Omega}}_1]\cup[-{\boldsymbol{\Omega}}_1,{-\boldsymbol{\Omega}}_0]$, similar to the one-dimensional case. This is a key requirement, as the energy of nonlinear functions is typically concentrated in finite domains.
\begin{theorem}\label{thm1}
Let the orthogonal wavelet functions
${\psi }_{mn}\left(\boldsymbol{x} \right)=2^{\frac{1}{2}md}\psi \left(2^{m}\boldsymbol{x}-n\right)$ with $m\in \textbf{Z}$ and $n\in \textbf{Z}^d$
form an orthonormal wavelet frame. Assume that Assumptions \ref{assum1}-\ref{assum3} hold for the nonlinear mapping $f(\cdot)\in L^2(\textbf{R}^d)$. Given $\varepsilon>0$, there exists a $B_\varepsilon$ defined as follows:
\bea
B_\varepsilon=B_{\varepsilon
}(\boldsymbol{T},{\boldsymbol{\Omega }}_0,{\boldsymbol{\Omega }}_1)\hspace{2.0in}\nonumber\\
=\{(m,n)|m_1 < m < m_0, |n|_s \preceq 2^{m} \boldsymbol{T} + \boldsymbol{t}_{\varepsilon} ,\boldsymbol{T}, \boldsymbol{t}_{\varepsilon} \in \textbf{Z}^d\},
\label{definition_B_epsilon}
\eea
where $m_0,m_1, \boldsymbol{t}_{\varepsilon}$ will be specified in the proof. For any $\varepsilon >0$, the following inequality holds:
\begin{align}
&\left\|{f-\sum_{\left(m,n\right)\in B_{\varepsilon }}{\left\langle {\psi }_{mn},f\right\rangle {\psi }_{mn}}}\right\|\leq \|\left(\boldsymbol{1}-{\mathcal{P}}_{{\boldsymbol{\Omega }}_1}+{\mathcal{P}}_{{\boldsymbol{\Omega }}_0}\right)\widehat{f}\|\nonumber\\
& + \|\left(\boldsymbol{1}-{\mathcal{Q}}_{\boldsymbol{T}}\right)f\|+2{\left({2\pi }\right)}^{\frac{d}{2}}\varepsilon \| f\|.\label{eq15}
\end{align}
\end{theorem}

{\em Proof}: See the Supplementary Material. 
\hfill $\Box$

The next corollary shows that many wavelet coefficients are close to zero, meaning only a subset of wavelets significantly contribute to the approximation. This is crucial for expanding the wavelet bases in the proposed mechanism.
\begin{corollary}\label{coro1}
Assume that Assumptions \ref{assum1}-\ref{assum3} hold for the nonlinear mapping $f(\cdot)\in L^2(\textbf{R}^d)$. 
For a given approximation accuracy $\varepsilon$ and $B_{\varepsilon }$ as defined in (\ref{definition_B_epsilon}), orthogonal wavelets ${\psi }_{mn}(\boldsymbol{x})$ with $(m,n)\notin B_{\varepsilon }$ satisfy  
\begin{align}
\label{eq21}
\left|\left\langle{\psi }_{mn},f\right\rangle\right|\approx 0,\quad \left(\left(m,n\right)\notin B_{\varepsilon }\right).
\end{align}
\end{corollary}

{\em Proof}: See the Supplementary Material. 
\hfill $\Box$

We will identify the most effective wavelets for approximation and design a mechanism to expand the wavelet bases.

\subsection{An algorithm to increase the wavelet bases}

The wavelet-basis increase algorithm aims to achieve desired approximation accuracy with a minimal number of wavelets, reducing computational cost. Since wavelets are orthogonal, adding more does not affect previously trained coefficients. Additionally, based on Corollary \ref{coro1}, not all bases in the new subspace $W_{m+1}$ need to be added. The key challenges are selecting appropriate bases in $W_{m+1}$ and determining when to stop the process.

Next, we describe the process for determining the minimal number of bases to achieve approximation accuracy $\varepsilon$ for a dataset $\{\boldsymbol{x}_i, y(\boldsymbol{x}_i)\}$. This involves iteratively increasing the number of bases until the desired accuracy is reached. The goal is to identify wavelet bases that closely match $f(\cdot)$ and select appropriate bases in $W_{m+1}$. The algorithm takes as input the initial bases from $W_{m_{init}}$, partition parameter $\mu$, and target accuracy $\varepsilon$, then outputs the number of iterations $k$ and the wavelet translation centers (WTC).

\begin{figure}[!t]
\centering
\includegraphics[width=0.35\textwidth]{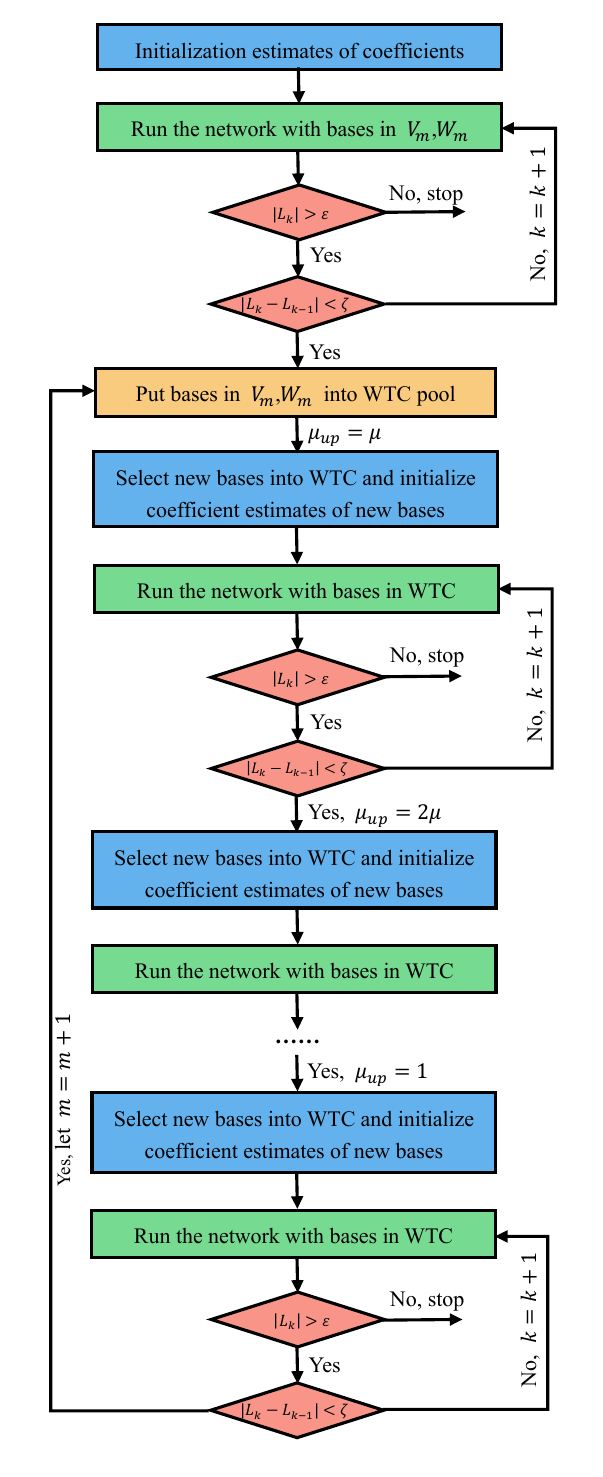}
\caption{Flowchart of the wavelet-basis increase algorithm}
\label{fig5}
\end{figure}

Algorithm \ref{alg2} is presented, with its corresponding flowchart in Fig.\,\ref{fig5}.

\begin{algorithm}[!htb]
\label{alg2}
\SetAlgoLined
\caption{Wavelet-basis increase algorithm}
\KwIn{The translation centers ${\boldsymbol{K}}^{\boldsymbol{m_{init}}}_{in}$, the training data $\{\boldsymbol{x}_i,y{(\boldsymbol{x}_i)}\}_{i=1,2,\dots,N_{sa}}$, the termination threshold $\zeta$, parameter $\mu$, the approximation accuracy $\varepsilon$.}
\KwOut{Iteration number $k$, basis functions in WTC.}
${m}\leftarrow m_{init}$\;
$k\leftarrow 1$\;
Initialize the estimates of all coefficients $\widehat{\theta}_{j,0}$ to be $0$\;
Using all bases in $V_m$ and $W_{m}$, run the neural network to estimate the coefficients $\widehat{\theta}_{j,1}$\;
\While{$|{L }_k-L_{k-1}|>\zeta$}{
\eIf{$|L_k| >\varepsilon$}{Using all bases in $V_m$ and $W_{m}$, run the neural network to estimate the coefficients $\widehat{\theta}_{j,k+1}$\;
Compute the approximation error $L_{k+1}$ using \eqref{Lk}\;}{break\;}
$k\leftarrow k+1$\;
}

{\bf output}

Put all bases in $V_m$ and $W_{m}$ into the WTC pool\;
\While{the approximation accuracy $\varepsilon$ is not achieved}{
${\mu }_{up}={\mu}$\;
\While{ {${\mu }_{up}\le 1$ }}{
Select translation centers ${\boldsymbol{K}}^{\boldsymbol{m}}$ in $W_{m}$ based on the parameter ${\mu }_{up}$ and \eqref{eq22}\;
Collect basis functions with translation centers ${\boldsymbol{K}}^{\boldsymbol{m+1}}$ into the WTC pool using \eqref{eq23}\;
Initialize the coefficient estimates of the newly selected bases to be $0$\;
\While{$|{L}_k-L_{k-1}|>{\zeta}$}{
\eIf{$|{L}_k|>\varepsilon$}{Run the neural network using the bases in the WTC pool to estimate the coefficients $\widehat{\theta}_{j,k+1}$\;
Compute the approximation error $L_{k+1}$ using \eqref{Lk}\; }{Stop the whole procedure\;}
$k \leftarrow k+1$\;
}

${\mu }_{up}\leftarrow{\mu }_{up}+\mu $\;}
$m\leftarrow m+1$\;}
\end{algorithm}

\emph{Step 1:} Determine the wavelet bases used to approximate $f(\cdot)$, denoted as WTC.

It begins with the initial resolution $m_{init}$ from Algorithm \ref{alg1}. Let $m = m_{init}$, and place the bases in space $W_m$ with translation centers ${\boldsymbol{K}}^{\boldsymbol{m}}_{in} = \{[K^{m,i_{m}}_1, K^{m,i_{m}}_2, \dots, K^{m,i_{m}}_d]\}$ into WTC, where $i_m = 1, 2, \dots, N_m$, and $N_m$ is the number of bases. Then, approximate the nonlinear function in \eqref{syst_1} iteratively using all bases in the scaling subspace $V_m$ and wavelet subspace $W_m$, with translation centers ${\boldsymbol{K}}^{\boldsymbol{m}}_{in}$ and data points $\{\boldsymbol{x}_i, y(\boldsymbol{x}_i)\}_{i=1,\ldots,N_{sa}}$. Note that $V_m$ is orthogonal to $W_m$ and approximates the low-frequency part of $f(\cdot)$; see \cite{21} for more details.

At the $k^{th}$ iteration, we denote the approximation error as
\begin{align}\label{Lk}
L_k:= \frac{1}{N_{sa}}\sum^{N_{sa}}_{i=1}(y(\boldsymbol{x}_i)-\widehat{y}_k(\boldsymbol{x}_i))^2,
\end{align}
where $\widehat{y}_k$ is the current step estimation using WNN, i.e.,
$\widehat{y}_k(\boldsymbol{x}_i)=\sum_{j=1}^{N_m}{\widehat{\theta }_{j,k}{\psi }_j(\boldsymbol{x}_i) }$,
where $\widehat{\theta }_{j,k}$ denotes the estimation of $\theta_j$ for the $j$th wavelet basis at the $k$th iteration. All $\widehat{\theta }_{j,0}$ are initialized to $0$, and the process continues until the termination threshold ${\zeta}$ is reached.

After computing the wavelet bases in $W_{m}$, we partition them into segments using energy estimation $\widehat{E}_m$, obtained from \eqref{eq10} with updated basis coefficients.

Let $\mu \in (0,1)$ be the user-defined separation factor, which is a reciprocal of a positive integer. We first select $N_{\mu \cdot \widehat{E}_{m}}$ wavelets from $W_m$ with the highest energy, where $N_{\mu \cdot \widehat{E}_{m}}$ is the number of bases whose energy is at least $\mu \cdot \widehat{E}_{m}$, rounded up to the nearest integer. The translation centers of these wavelet bases are then defined as
\begin{align}
\label{eq22}
{\boldsymbol{K}}^{\boldsymbol{m}}=\left\{\left[K^{m,i_{m}}_1,\ K^{m,i_{m}}_2,\dots,K^{m,i_{m}}_d\right]\right\},
\end{align}
where $i_{m}=1,2,\dots, N_{\mu\cdot E_{m}}$, and $K^{m,i_m}_l$ denotes the translation center of the $l^{th}$ dimension, $l= 1,2,...,d$.

\emph{Step 2:} The wavelet bases in $W_{m+1}$ with translation centers closest to ${\boldsymbol{K^{m}}}$ are selected, with translation centers denoted as ${\boldsymbol{K^{m+1}}}$. Let ${\boldsymbol{K^{m+1}}_l}=\{K^{{m+1},1}_l$, $K^{{m+1},2}_l$, $\dots$, $K^{{m+1}, {(N{m+1})}^{\frac{1}{d}}}_l\}$ represent the candidate translation centers for the $l^{th}$ dimension, where $l=1,2,\dots,d$. We obtain
\begin{align}
&{\boldsymbol{K}}^{\boldsymbol{{m+1}}}=\left\{[K^{{m+1},i_{m+1}}_1,\ K^{{m+1},i_{m+1}}_2,\dots,K^{{m+1},i_{m+1}}_d] \right\}, \nonumber\\
&K^{{m+1},i_{m+1}}_l\in \mathscr{U}( K^{m,i_{m}}_l,{\boldsymbol{K}}^{\boldsymbol{{m+1}}}_{l}),\label{eq23}
\end{align}
where $i_{m+1} = 1,2,...,2^d\cdot N_{\mu*\widehat{E}_{m}}$, $\mathscr{U}( K^{m,i_{m}}_l,{\boldsymbol{K}}^{\boldsymbol{{m+1}}}_{l})$ has two elements, and $K^{{m+1},i_{m+1}}_l$ fetch through the elements of $\mathscr{U}( K^{m,i_{m}}_l,{\boldsymbol{K}}^{\boldsymbol{{m+1}}}_{l})$.

\emph{Step 3:} All basis functions in $V_m$, $W_m$, and the newly selected bases in $W_{m+1}$ are stored in a set WTC, initially empty. Coefficients of the newly selected bases are initialized to zero. A neural network is executed using the basis functions in WTC until the termination threshold $\zeta$ is reached. Here, $\zeta$ represents the stopping condition.
Following this, $N_{2\mu \cdot \widehat{E}_{m}}$ bases with the largest energy are selected based on energy contribution $2\mu \cdot \widehat{E}_{m}$. Previously selected $N_{\mu \cdot \widehat{E}_{m}}$ bases are not reselected. New wavelet bases are added to WTC according to \eqref{eq23}, ensuring no duplicate elements. The process repeats until the algorithm achieves the specified approximation accuracy $\varepsilon$. If $\varepsilon$ is not met after adding all bases in $W_{m+1}$, $m$ is incremented by 1, and the algorithm restarts from \emph{Step 1}.

\begin{remark}
The performance of the approximation depends on the separation factor $\mu$. A smaller $\mu$ results in a finer partitioning of the wavelet subspace, allowing better utilization of wavelet bases but potentially slowing convergence speed in real-time applications. Conversely, a larger $\mu$ increases computational resource consumption. Selecting an appropriate value of $\mu$ is case-dependent, especially since the nonlinear mapping $f(\cdot)$ is unknown. In our simulations (Example 1), four values of $\mu$, ranging from $\frac{1}{5}$ to $\frac{1}{2}$, are used to illustrate how these values affect the number of parameters or wavelets required for approximation under three different accuracy requirements. This example provides some insight into how to tune this parameter.
\end{remark}
\begin{remark}
The termination threshold, denoted as $\mathcal{TH}$, determines the timing for introducing new basis functions.
A large termination threshold may prematurely trigger the addition of new basis functions, potentially
resulting in insufficient training of existing parameters and the introduction of redundant parameters.
Conversely, an excessively small termination threshold may induce network overfitting and impede
loss convergence speed. Generally, the termination threshold should scale inversely with the monitoring accuracy threshold $\varepsilon$, requiring proportional reduction as $\varepsilon$ decreases. Based on empirical
observations, we recommend setting the termination threshold within the range of $0.01\varepsilon$ to $0.001\varepsilon$
for optimal performance.
\end{remark}

\begin{remark}
The primary objective of approximating $f(\cdot)$ using the subspace $W_{m}$ and its bases is to find wavelet bases that closely match $f(\cdot)$'s waveform and then select appropriate bases in $W_{{m+1}}$. If $W_{m}$ bases significantly reduce approximation error, adding new bases in $W_{{m+1}}$ might slow parameter updates and hinder convergence. To avoid this, ${\zeta}$ in Algorithm \ref{alg2} is set relatively large. In \cite{25}, achieving tracking accuracy in $W_{m}$ required many iterations and strict thresholds. As the tracking error decreases, the threshold tightens. When $\mu$ is fixed, ${\zeta}$ in Algorithm \ref{alg2} should be small. Typically, $\zeta$ is chosen to be 10 times smaller than the desired accuracy, depending on the application.
\end{remark}

\begin{remark}
Analyzing computational complexity is essential for any functional approximation technique, though it can be challenging. Dimensionality causes exponential growth in cost, expressed as $P(d)=O(n^d)$, where $n$ is the number of parameters per dimension and $d$ is fixed by the input variables. Thus, controlling the number of parameters per dimension is key to reducing complexity. This paper’s main idea is to estimate the spatial frequency of the unknown function to substantially reduce parameters and computational cost. As shown in Example 1, the traditional wavelet neural network requires $420$ parameters to reach the desired accuracy, while our method achieves similar results with only $174$ parameters. On the other hand, the size of dataset also affects the computational cost. For large datasets, time complexity typically scales linearly with batch size during iterative training: larger batches increase per-iteration cost but improve gradient estimates, potentially lowering total iterations. Storage complexity also grows with batch size due to the need to store training samples and intermediate states. These factors highlight trade-offs between algorithm design and resource constraints in high-dimensional, data-intensive tasks. Our approach significantly reduces parameter-related costs, though training time and storage are also affected by dataset size.
\end{remark}

\begin{remark}
Although this paper assumes the nonlinear mapping $f (\cdot)$ has a stationary frequency
distribution in the spatial domain, the proposed method can also handle non-stationary frequency
distributions, as it can add wavelet bases with high-frequency components when the desired approximation
accuracy is not achieved. While a detailed analysis is not included in this paper, the proposed method
can also remove unnecessary low-frequency components if their weights become sufficiently small,
providing flexibility in capturing spatial frequency variations. In our simulation example, Example 3 shows that the proposed CWNN is not sensitive to the frequency shift.
\end{remark}

\section{Applications}\label{sec5}

This section demonstrates the adaptability of the proposed constructive wavelet neural network (CWNN) in approximating unknown mappings through various examples. Unlike traditional fixed-architecture NNs, CWNN dynamically adjusts its structure to achieve the desired accuracy, making it well-suited for both offline and online tasks such as time-series prediction.

Four examples are presented to demonstrate the effectiveness of the proposed method. Example 1 illustrates its ability to learn a nonlinear mapping from an offline dataset, providing insights into parameter selection and evaluating robustness under both small and large noise levels. Example 2 highlights the method’s effectiveness in learning from two separate offline datasets, each containing partial spatial information. Example 3 showcases its capability to handle time-series data in which the nonlinear mapping changes at specific time instants. Example 4 validates the method’s nonlinear learning capacity using real-world time-series datasets.

Examples 1 and 4 include direct comparisons with the standard wavelet neural network (WNN). While both CWNN and WNN can add new bases to improve approximation accuracy, their structure adjustment mechanisms differ significantly. WNN randomly selects initial wavelet bases and adds new ones based solely on approximation error. In contrast, CWNN dynamically adjusts its structure in response to data characteristics, enabling more efficient and targeted learning.

To benchmark against graph neural networks (GNNs), three fixed-architecture variants—GNN-3, GNN-5, and GNN-7—are employed. Each variant consists of multi-layer perceptrons (MLPs) with ReLU activations, residual connections, adaptive initialization, and batch normalization. These networks feature 3, 5, or 7 layers, each with 128 hidden nodes, capturing a range of model complexities to ensure a robust baseline for comparison. For fairness, the weights of all five neural networks are updated using standard backpropagation.

\noindent \underline{Example 1}: Learning a Mapping from an Off-line Data Set

Example 1 demonstrates how to tune the parameters of the proposed method and evaluates its performance—considering accuracy, generalizability, computational cost, and robustness to measurement noise—in comparison with three GNN variants and the standard WNN.

The following two-dimensional mapping is considered:
\begin{align}\label{syst_2}
y = 0.5 + x_1 +x_2 + \sin(2\pi(x_1 + x_2)) + d_1(x_1),
\end{align}
where $x_2 = \sqrt{x_1}$, $\boldsymbol{x} = [x_1,x_2],\  x_1,x_2\in[0,1]$. 

\begin{figure}[htb!]
\centering
\subfigure[]{
\includegraphics[width=0.30\textwidth]{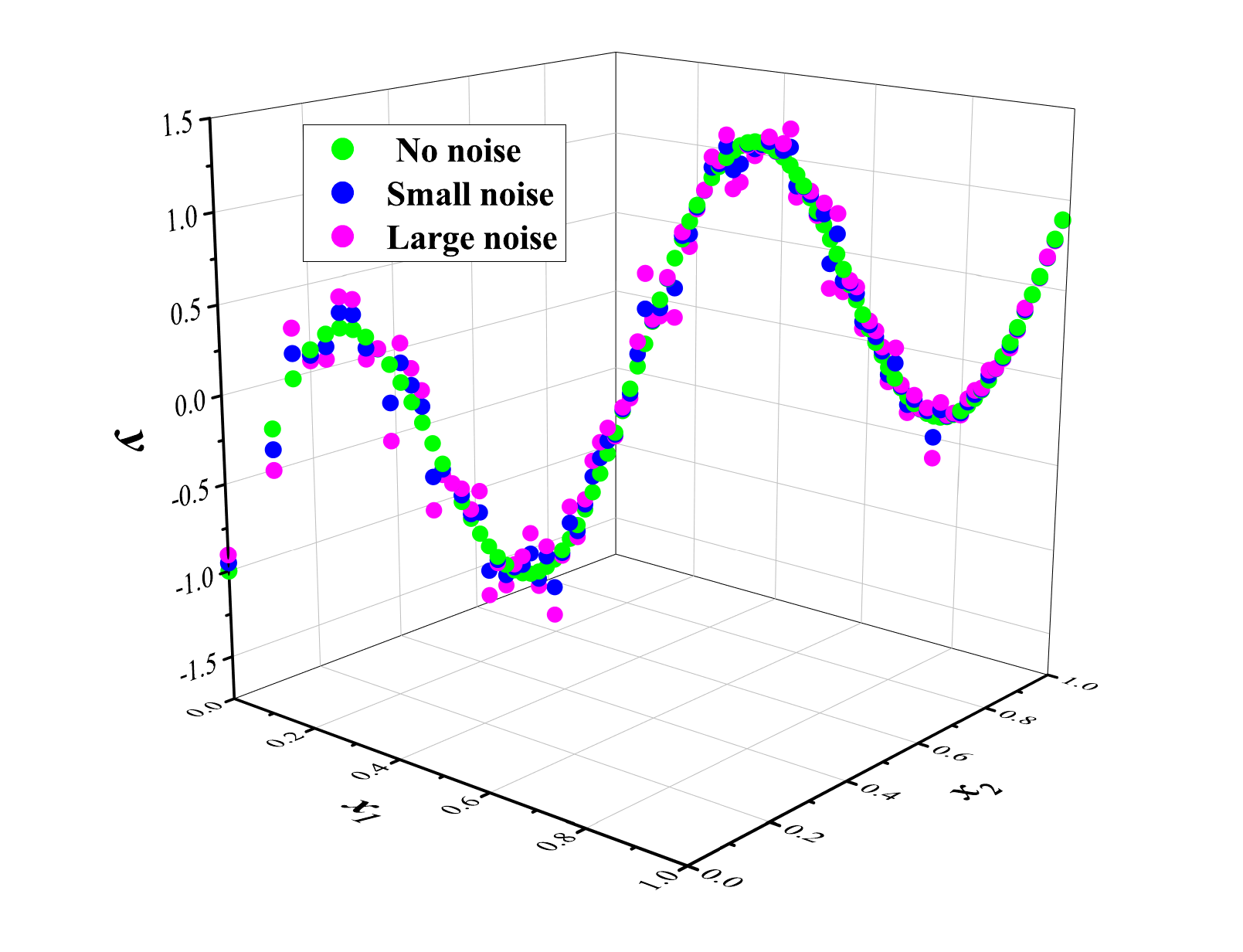}
\label{fignew7a}}
\subfigure[]{
\includegraphics[width=0.30\textwidth]{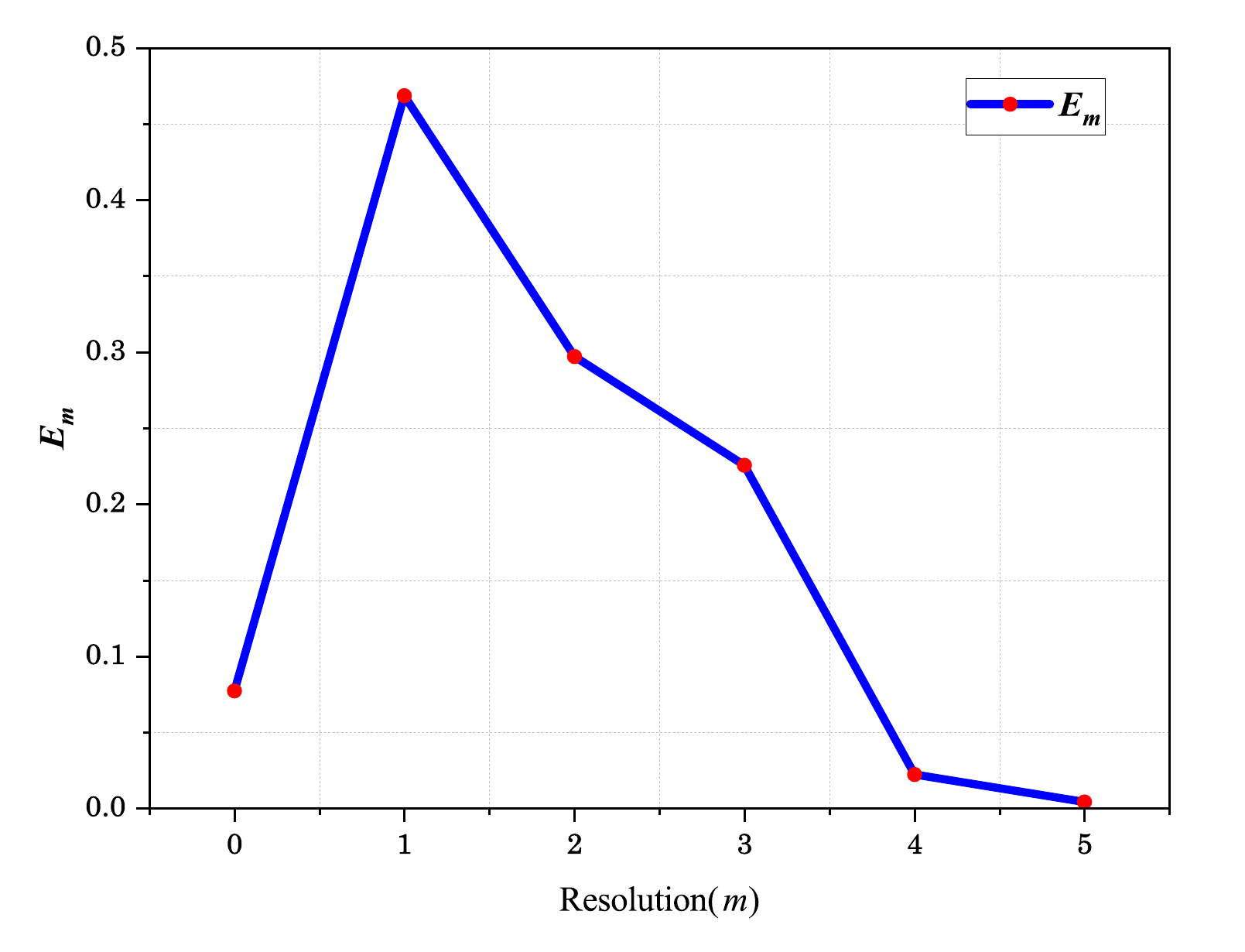}
\label{fignew7b}}
\centering
\caption{(a) $D_1$, $D_2$ and $D_3$. (b) Energy $E_m$ in subspace $W_m$.}
\label{fignew7}
\end{figure}

Three datasets are used in this example. The first dataset, \(D_1\), is generated from (\ref{syst_2}) with \(d_1(x_1) = 0\). This dataset is used to illustrate the selection of the tuning parameter \(\mu\) in the proposed method. Two additional datasets, denoted as \( D_2 \) and \( D_3 \), are generated by introducing noise to the original function. Specifically, in \( D_2 \), the output \( d_1(x_1) \) follows a normal distribution with zero mean and input-dependent variance: 
\[
d_1(x_1) \sim \mathcal{N}\left(0,\; \sigma^2(x_1)\right), \quad \text{where } \sigma(x_1) = 0.1(1 - x_1^2).
\]
For $D_3$, the noises $d_1$ also follow a normal distribution with zero mean and input-dependent variance. In this case, the noise variance is increased to:
\[
\sigma(x_1) = 0.2(1 - x_1^2).
\]
Figure~\ref{fignew7a} illustrates the three datasets.

In all datasets, \(x_1\) is uniformly sampled from \([0,1]\), and \(x_2 = \sqrt{x_1}\). Each dataset \(D_j, \ j = 1, 2, 3\), is split into training (\(D_{j,\text{train}}\)) and test (\(D_{j,\text{test}}\)) sets, with 80\% of the data randomly and uniformly allocated to the training set.

Sinc wavelet is used in the simulation. The one-dimensional Sinc wavelet is $2\sin(2x)/(2x)-\sin(x)/x$, satisfying conditions \eqref{neweq1}-\eqref{eq5}. Using \eqref{eq3}, the high-dimensional Sinc wavelet is derived. Fig.\,\ref{fignew7b} shows the energy $E_m$ projected onto the wavelet subspace $W_m$ via mapping \eqref{syst_2}. It exhibits unimodal behavior over $m$, consistent with Assumption \ref{assum2}.

\begin{figure*}[!t]
\centering
\subfigure[]{\label{fignew3a}
\includegraphics[width=0.32\textwidth]{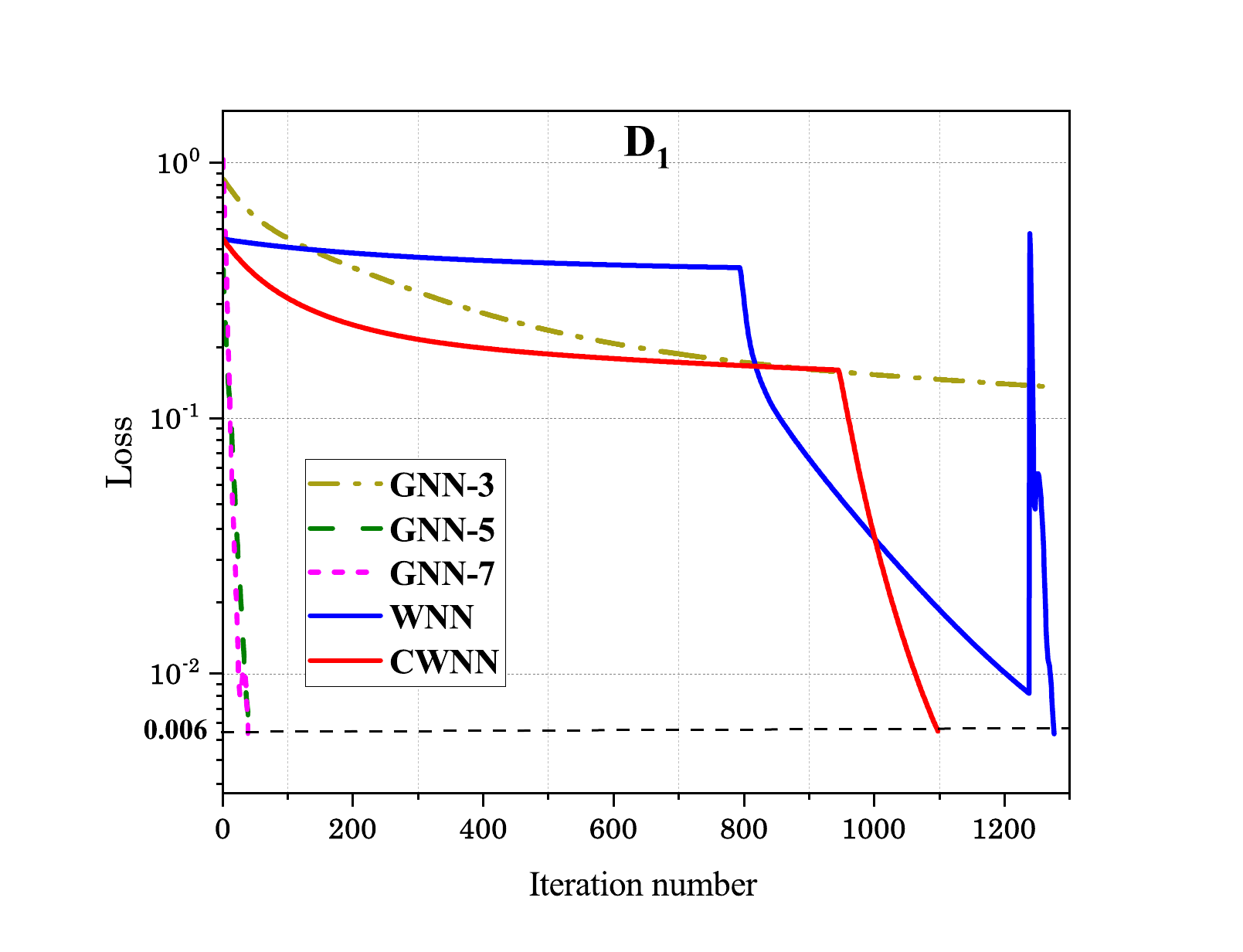}}
\subfigure[]{\label{fignew3b}
\includegraphics[width=0.32\textwidth]{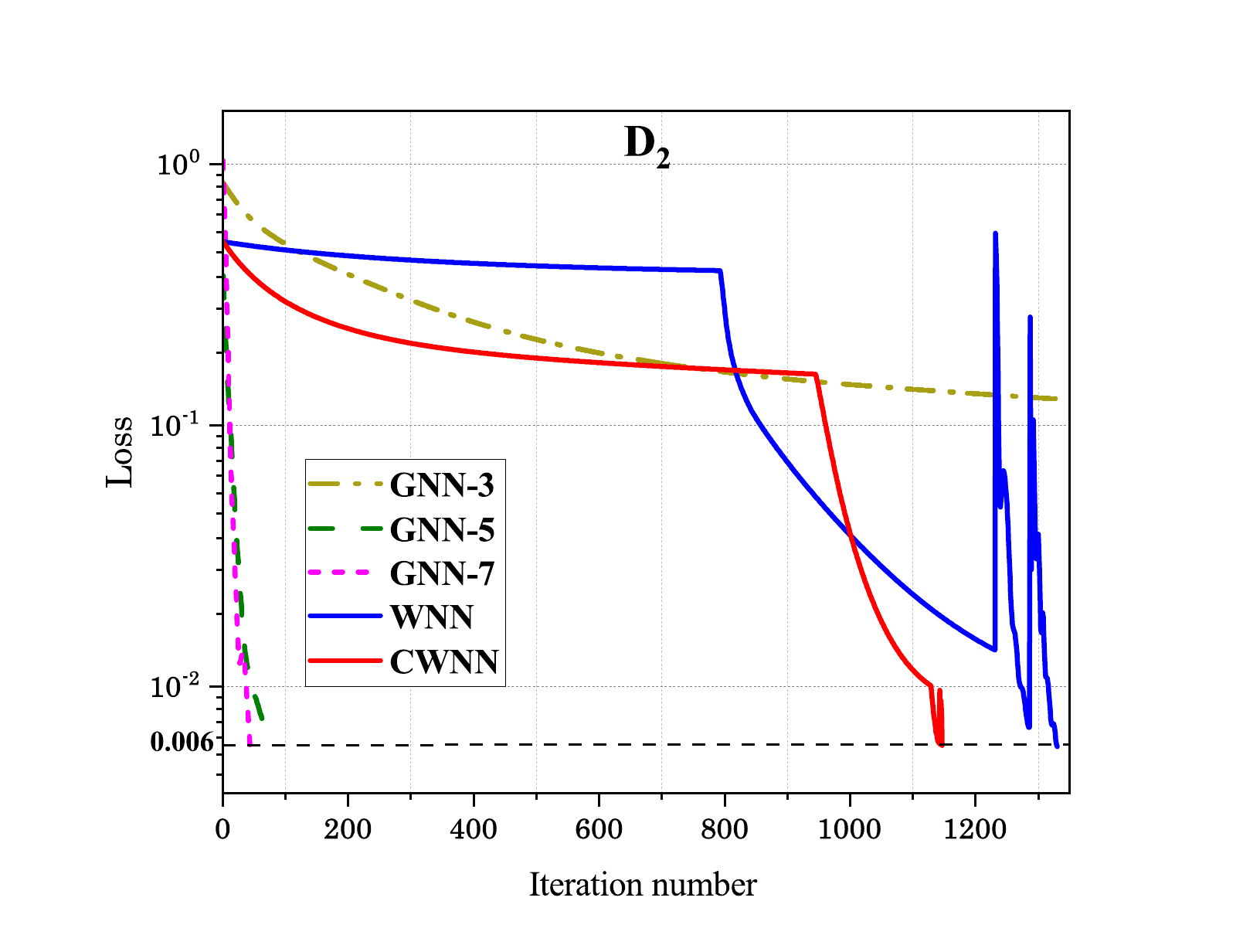} }
\subfigure[]{\label{fignew3c}
\includegraphics[width=0.32\textwidth]{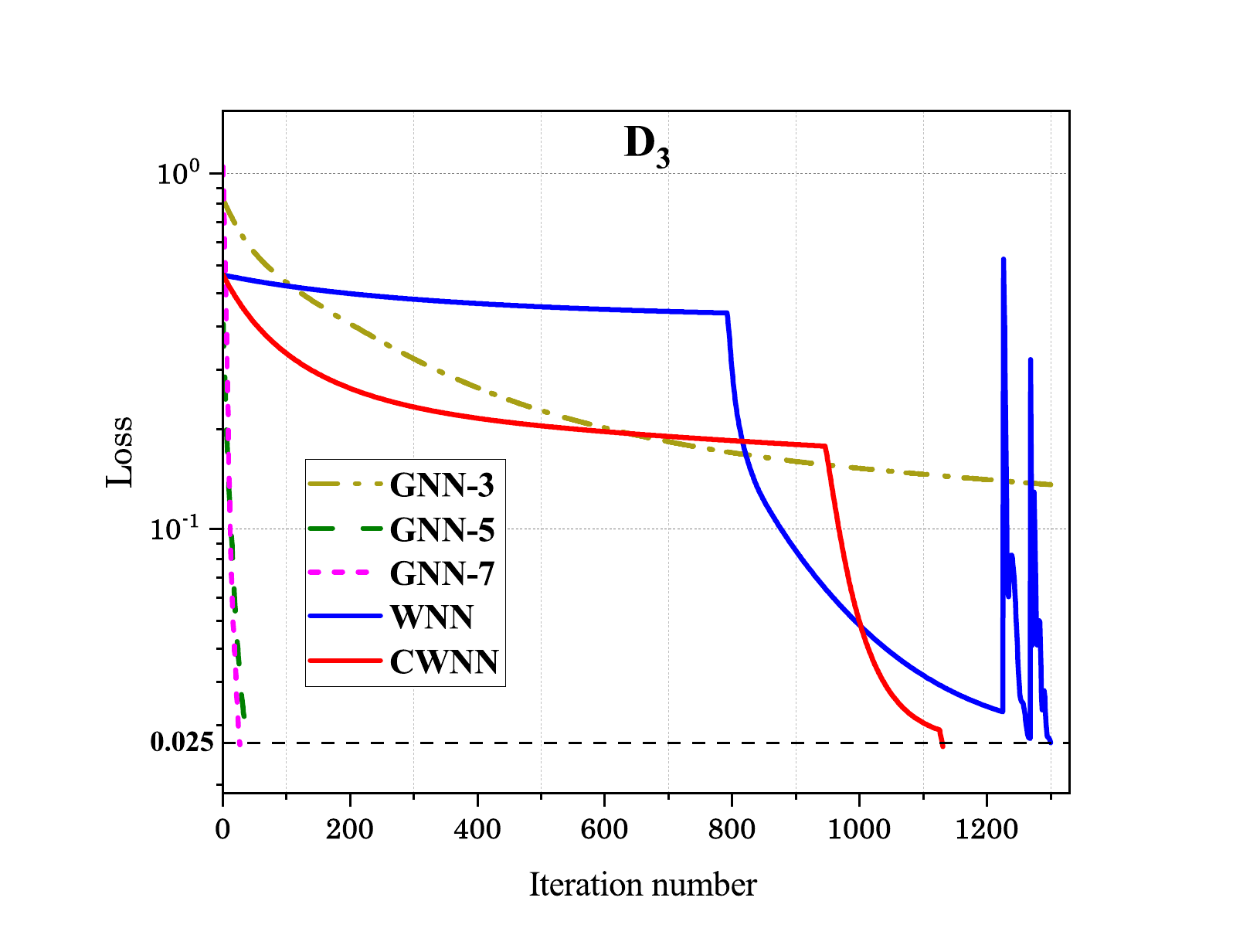}}
\caption{ The performance of training procedure using 5 neural networks and dataset $D_{1,\text{train}}, D_{2,\text{train}}, D_{3,\text{train}}$ respectively.}
\label{fignew3}
\end{figure*}

\noindent\emph{A: The Role of $\mu$}

The separation factor $\mu$ in Algorithm \ref{alg2} determines the selectivity of high-frequency subspace basis functions, resulting in variations in parameter requirements. Using the data set $D_{1,\text{train}}$, we evaluate this relationship by counting the number of parameters until the performance convergence to thresholds $\varepsilon = 0.0025, 0.01, 0.02$, as shown in Table \ref{tablenew2}. 


Larger separation factors, such as \(\mu = \frac{1}{2}\) and \(\mu = \frac{1}{3}\), generally require more basis functions than smaller ones like \(\mu = \frac{1}{4}\) or \(\mu = \frac{1}{5}\). This trend is particularly evident under stricter precision requirements. For instance, when \(\varepsilon\) decreases from 0.02 to 0.0025, the \(\mu = \frac{1}{3}\) configuration requires 43.7\% more parameters (174 \(\rightarrow\) 250), while the \(\mu = \frac{1}{5}\) setup sees a 27.7\% increase (166 \(\rightarrow\) 212). However, the \(\mu = \frac{1}{2}\) configuration keeps 190 bases for the three thresholds. This suggests that as \(\varepsilon\) decreases, the relationship between the separation factor and the required number of basis functions becomes more complex, highlighting the need for a careful balance in parameter selection.

\begin{table}[!htb]
\renewcommand{\arraystretch}{1.5}
\caption{ The number of parameters needed to reach the desired approximation accuracy -- Impact of $\mu$ }
\label{tablenew2}
\begin{tabular*}{0.5\textwidth}{p{70pt}|p{50pt}|p{50pt}|p{50pt}}
\hline
\textbf{separation factor $\mu$} &
\textbf{$\varepsilon=0.0025$} &
\textbf{$\varepsilon=0.01$} &
\textbf{$\varepsilon=0.02$}\\
\hline
\textbf{$\mu$ = 1/2} & \textbf{190} & 190 & 190\\
\hline
\textbf{$\mu$ = 1/3} & 250 & 174 & 174\\
\hline
\textbf{$\mu$ = 1/4} & 228 & 170 &  170\\
\hline
\textbf{$\mu$ = 1/5} & 212 & \textbf{166} &  \textbf{166}\\
\hline
\end{tabular*}
\end{table}

\noindent\emph{B: Performance Comparison} 

We compare the performance of the proposed method (CWNN) with WNN, GNN-3, GNN-5, and GNN-7 in terms of accuracy using the traning sets and testing tests respectively,  computational cost, and robustness. As discussed in Remark 7 computational cost is measured by the number of parameters required during training. Robustness is evaluated using the noisy datasets \(D_2\) and \(D_3\), where the trained networks are tested on the corresponding testing sets \(D_{2,\text{test}}\) and \(D_{3,\text{test}}\).

In this comparison, the parameter $\kappa$ of CWNN is set to $0.36$, balancing computation resources and application requirements. The learning rate $\iota_r$ is set to $0.0005$. Initially, the frequency is calculated using Algorithm \ref{alg1}. Since $\boldsymbol{x}$ in \eqref{syst_2} is bounded by $[0,1]\times[0,1]$, the translation centers $\{{\boldsymbol{K}}_{1,1},{\boldsymbol{K}}_{1,2},\dots,{\boldsymbol{K}}_{{1,N_1}}\}$ are given as $\{ 0,0.5,1,1.5,2 \} \times \{ 0,0.5,1,1.5,2 \}$. With $\kappa=0.36$, $\{{\boldsymbol{K}}_{1,1}$, ${\boldsymbol{K}}_{1,2}$, $\dots$, ${\boldsymbol{K}}_{{1,N}_{J}}\} = \{ 0,1,2 \} \times \{ 0,1,2 \}$. From Algorithm \ref{alg1}, the initial resolution $m_{init}$ is $2$, and the initial frequency is $2^2$. This value is then input into Algorithm \ref{alg2}, where the translation centers ${\boldsymbol{K}}^{\boldsymbol{m_{init}}}_{in}$ are given as $\{ 0,0.25,0.5,\dots,2 \} \times \{ 0,0.25,0.5,\dots,2 \}$. The parameter $\varepsilon$ is chosen as $0.006$ for datasets ${D_1}, {D_2}$, and $0.025$ for dataset ${D_3}$. Termination thresholds are all set to $0.00004$, and $\mu$ is set to $1/3$. Algorithm \ref{alg2} is subsequently applied.

Here the Loss is used to represents the approximation error for any given dataset $D$ with $N_{sa}$ data points. It is defined as 
\bea
\text{Loss} = \frac{1}{ N_{sa}} \sum^{N_{sa}}_{i=1}(y(\boldsymbol{x}_i) - \widehat{y}(\boldsymbol{x}_i))^2, \label{lost_function}
\eea
where $\boldsymbol{x}_i$ is the input, $y(\boldsymbol{x}_i)$ is the actual output, and $\widehat{y}(\boldsymbol{x}_i)$ is the neural network's output.

\noindent{\underline {Accuracy using training sets}} 

Fig.\,\ref{fignew3} shows the performance of the three methods using $Loss$ defined in (\ref{lost_function}) over iterations. For the three datasets \(D_{1,\text{train}}\), \(D_{2,\text{train}}\) and \(D_{3,\text{train}}\), it shows that GNN-5 and GNN-7 perform well and generally converge faster than CWNN and WNN, likely because they do not require structural adjustments. While GNN-3 fails to reach the desired approximation accuracy, both GNN-5 and GNN-7 succeed, indicating that increasing the number of layers typically enhances the approximation capability. However, determining the ``sufficient'' number of layers for a specific dataset and accuracy target often requires trial and error. For all datasets \(D_{1,\text{train}}\), \(D_{2,\text{train}}\) and \(D_{3,\text{train}}\), CWNN converges to the desired accuracy around \(1100^{\text{th}}\) iteration and WNN converges around the \(1300^{\text{th}}\) iteration, whereas GNN-3 does not attain the predefined accuracy within the given iterations.

\noindent{\underline{Accuracy using Testing sets}}

Table \ref{tablenew3} summarizes the accuracy of the five NNs across three test sets. Both WNN and CWNN maintain accuracy comparable to their training performance, indicating well-tuned parameters that avoid underfitting and overfitting. In contrast, the GNNs perform significantly worse, with substantially lower accuracy despite lower loss values. GNN-7, in particular, shows the poorest accuracy, suggesting overfitting due to excessive model complexity.

\begin{table}[htb!]
\renewcommand{\arraystretch}{1.5}
\caption{Accuracy with testing datasets }
\label{tablenew3}
\centering
\setlength{\tabcolsep}{3pt}
\begin{tabular}{p{35pt}|p{35pt}|p{35pt}|p{35pt}|p{35pt}|p{35pt}}
\hline
\textbf{Method} &
\textbf{ GNN-3} &
\textbf{ GNN-5} &
\textbf{ GNN-7}&
\textbf{ WNN}&
\textbf{ CWNN}\\
\hline 
\textbf{$\boldsymbol{\mathrm{D_1}}$} & $0.1417$ & $0.1090$ & $ 0.1696 $ & $\bf 0.0059$ & $0.0068$\\
\hline
\textbf{$\boldsymbol{\mathrm{D_2}}$} & $0.1607 $ & $0.0145 $ & $0.1652$ & $\bf 0.0047$ &  $0.0058$ \\
\hline
\textbf{$\boldsymbol{\mathrm{D_3}}$} & $0.1929 $ & $0.1428 $ & $0.2264$ & $0.0238$ &  $ \bf 0.0154$\\
\hline
\end{tabular}
\end{table}
\begin{remark}
It is important to note that selecting the optimal GNN architecture is not the focus of this work. While deeper GNNs generally offer better approximation accuracy, our use of different GNN variants highlights a key issue: although GNN-7 performs well on noise-free data, it struggles with noisy data. Its performance on all three test sets is the worst among the GNNs, indicating overfitting. This underscores a key challenge with GNNs—selecting an appropriate architecture to achieve reliable accuracy without either overfitting or underfitting.
\end{remark}

\noindent{\underline {Computational cost using three training sets}}

\begin{figure}[!htb]
\centering
\subfigure[]{\includegraphics[width=0.4\textwidth]{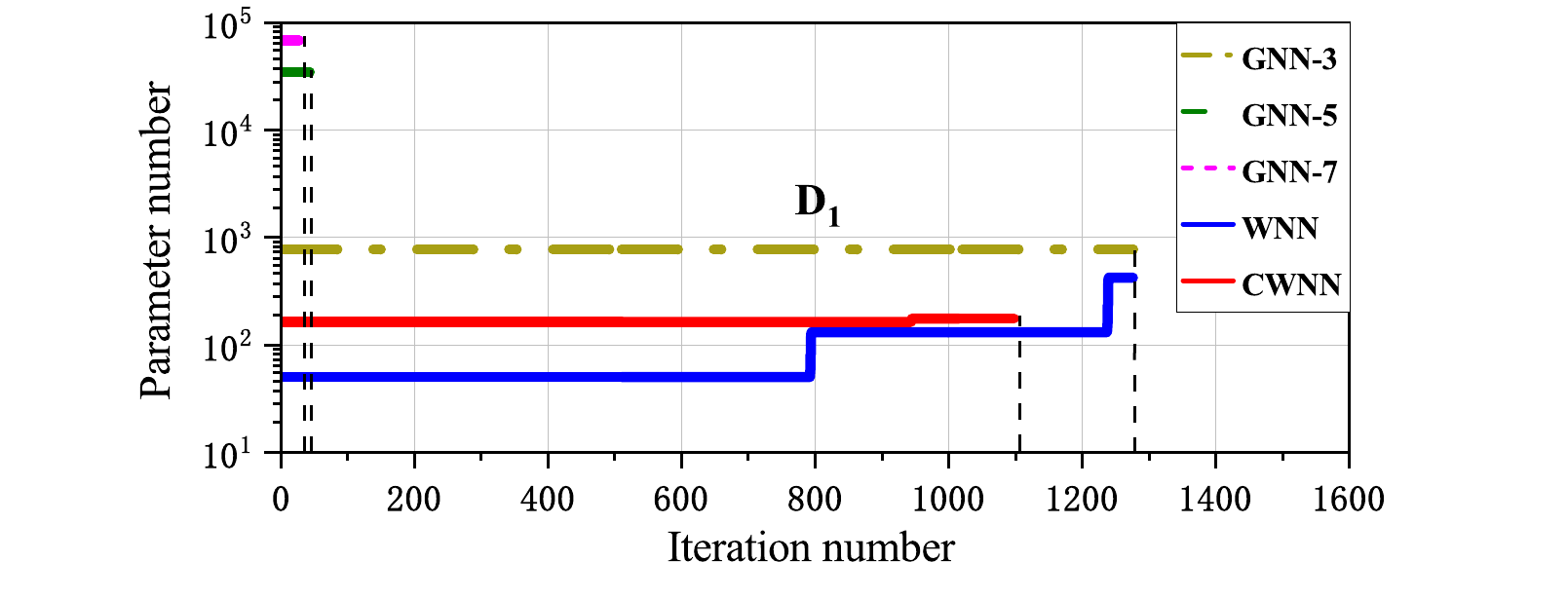}
\label{fignew8a}}
\subfigure[]{\includegraphics[width=0.4\textwidth]{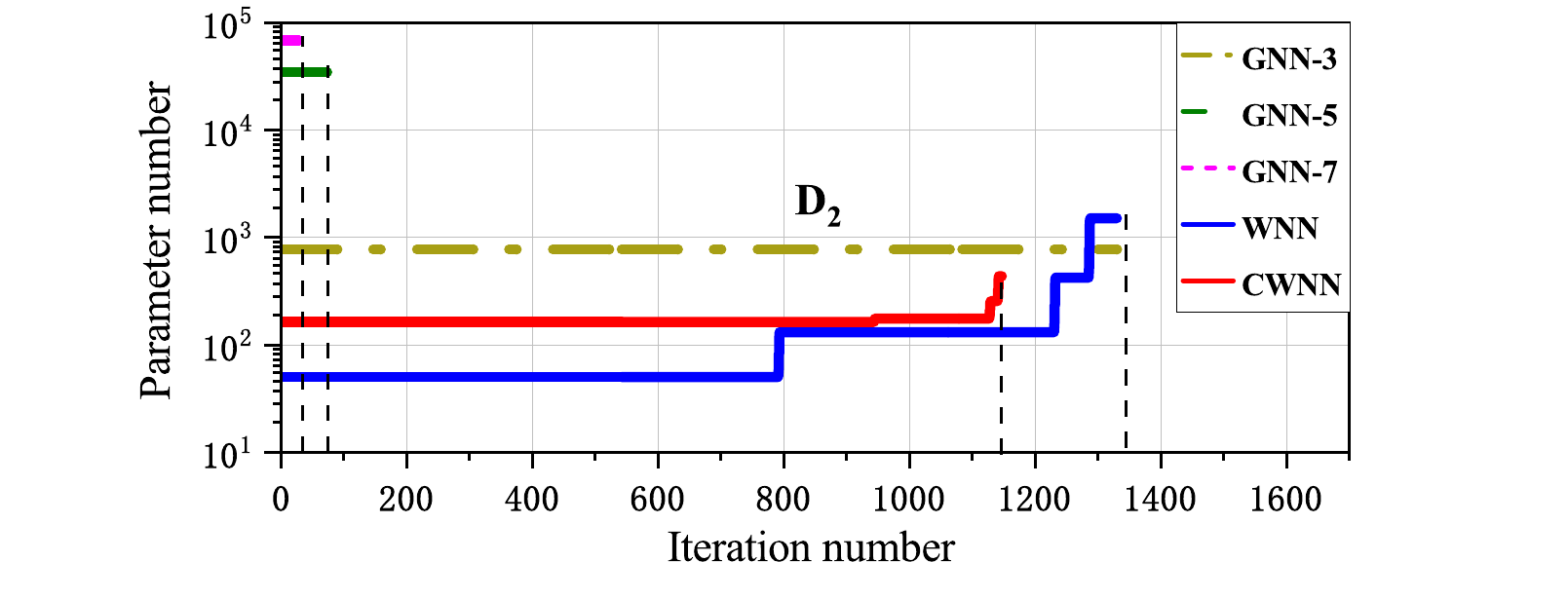}
\label{fignew8b}}
\subfigure[]{\includegraphics[width=0.4\textwidth]{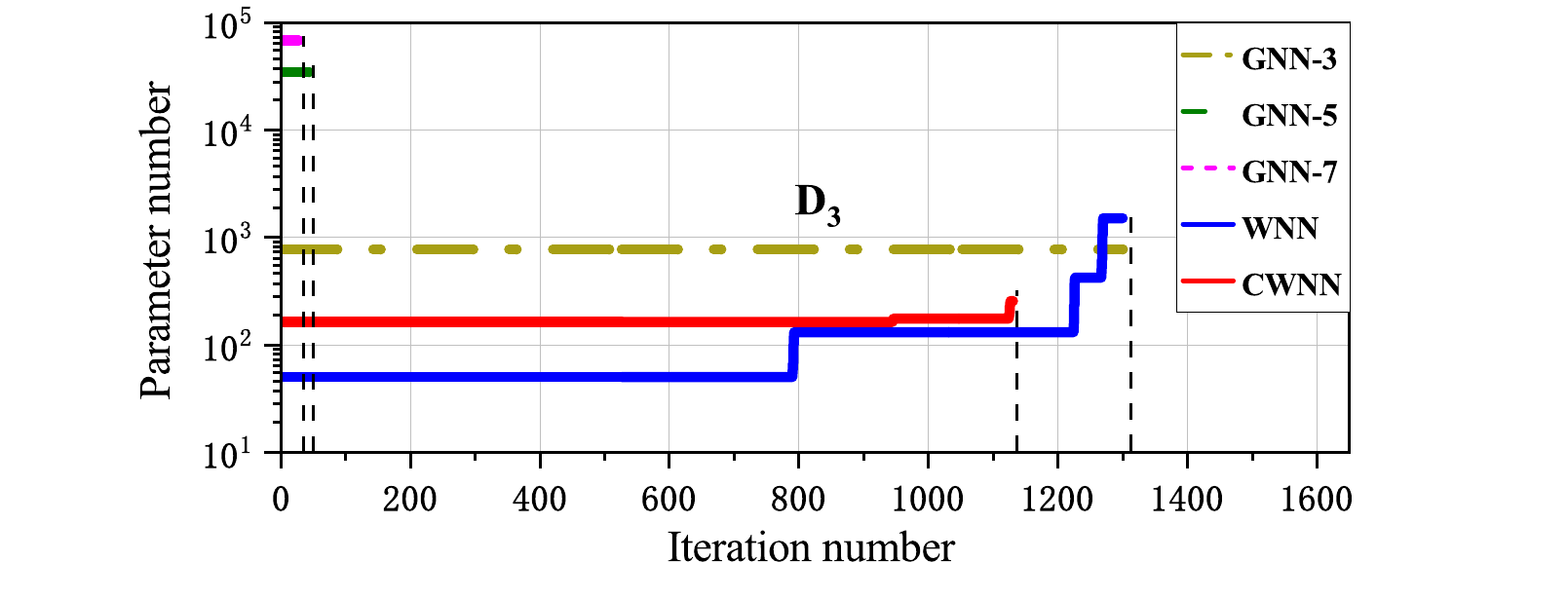}
\label{fignew8c}}
\caption{ The number of parameters used by 5 NNs using three training sets}
\label{fignew8}
\end{figure}

In these 5 NNs, parameters correspond to the bases in the hidden layer, as these architectures only optimize hidden node weights during learning. Fig.\,\ref{fignew8} shows the parameter variations for five networks. For $D_{1,\text{train}}$ dataset (see Fig.\,\ref{fignew8a}), GNN-3, GNN-5, and GNN-7 use 769, 34,305, and 67,841 parameters, respectively. WNN starts with 50 parameters in \(W_1\), adds 81 parameters in \(W_2\) at the \(794^{\text{th}}\) iteration, and reaches 420 parameters by the \(1239^{\text{th}}\) iteration. CWNN is similar, adding bases via Algorithms \ref{alg1} and \ref{alg2}, reaching 174 parameters by the \(945^{\text{th}}\) iteration. Clearly, GNN-based methods require far more parameters than CWNN and WNN, while CWNN achieves the desired accuracy with lower computational costs (a 58.6\% reduction compared with WNN).

For \(D_{2,\text{train}}\) and \(D_{3,\text{train}}\), the computational overhead varies across datasets, as shown in Figs.\,\ref{fignew8b} and \ref{fignew8c}. Results reveal distinct adaptation patterns. Under low noise (\(D_2\)), WNN expands to 1,509 bases by the \(1287^{\text{th}}\) iteration, while under high noise (\(D_3\)), it reaches 1,509 bases by the \(1309^{\text{th}}\) iteration. In contrast, CWNN demonstrates noise-robust convergence and parametric efficiency, stabilizing at 433 bases for \(D_2\) and 254 bases for \(D_3\), and achieving at least 3.5-fold reduction compared to WNN under high noise. GNN variants (GNN-3: 769, GNN-5: 34,305, GNN-7: 67,841 parameters) maintain fixed parameter counts regardless of noise intensity.

The performance across the three training sets shows that the proposed CWNN can achieve the desired approximation accuracy in the presence of measurement noises, with faster convergence speed and lower computational cost compared with WNN. 

\noindent{\underline {Robustness using two testing sets}}

Once the five NNs are trained, we use them to test performance on the testing sets. To evaluate the robustness of these networks, only \(D_{2,\text{test}}\) and \(D_{3,\text{test}}\) are used. To account for random noise effects, we generate 100 different $D_2$ and $D_3$ sets, producing 100 corresponding test sets. We evaluate the five trained NNs on these $D_{2,\text{test}}$ and $D_{3,\text{test}}$ sets, computing the average accuracy in terms of Loss and its variance. Table~\ref{tablenew1} summarizes the accuracy and standard deviation of the five methods. CWNN achieves similar accuracy to WNN but exhibits slightly less variation. Furthermore, GNN-3, GNN-5, and GNN-7 show lower accuracy compared to both CWNN and WNN, along with larger standard deviations. Both WNN and CWNN demonstrate robustness, as their performance is not sensitive to noise, indicating that dynamically adjusting the network structure enhances robustness.

\begin{table}[!htb]
\renewcommand{\arraystretch}{1.5}
\caption{Robustness of 5 NNs }
\label{tablenew1}
\centering
\setlength{\tabcolsep}{3pt}
\begin{tabular}{p{60pt}|p{80pt}|p{80pt}}
\hline
\textbf{Method} &
\textbf{$\boldsymbol{\mathrm{D_2}}$} &
\textbf{$\boldsymbol{\mathrm{D_3}}$}\\
\hline
\textbf{GNN-3} & $0.1638 \pm 0.1113$ & $0.1782 \pm 0.1172$\\
\hline
\textbf{GNN-5} & $0.0139 \pm 0.0217$ & $0.0295 \pm 0.0250$\\
\hline
\textbf{GNN-7} & $0.0217 \pm 0.0390$ & $0.0407 \pm 0.0445$\\
\hline
\textbf{WNN} & $0.0060\pm 0.0017$ & $0.0197\pm 0.0065$\\
\hline
\textbf{CWNN} & $0.0056\pm 0.0014$ &  $0.0207\pm 0.0063$\\
\hline
\end{tabular}
\end{table}

\noindent {\underline{Example 2}}: Learning a Mapping from Two Off-line Data Sets

In this example, we utilize the two-dimensional nonlinear mapping \eqref{syst_2} and collect two noise-free datasets in two regions of $A \times B$ with $x_1 \in A$ and $x_2 \in B$, namely, $DS_1: [0, 0.6] \times [0, \sqrt{0.6}]$ and $DS_2: [0.6, 1] \times [\sqrt{0.6}, 1]$. Initially, we uniformly sample $x_1$ from $[0, 1]$ and compute the output $y$ using \eqref{syst_2}.

\begin{figure}[!htb]
\centering
\includegraphics[width=0.35\textwidth]{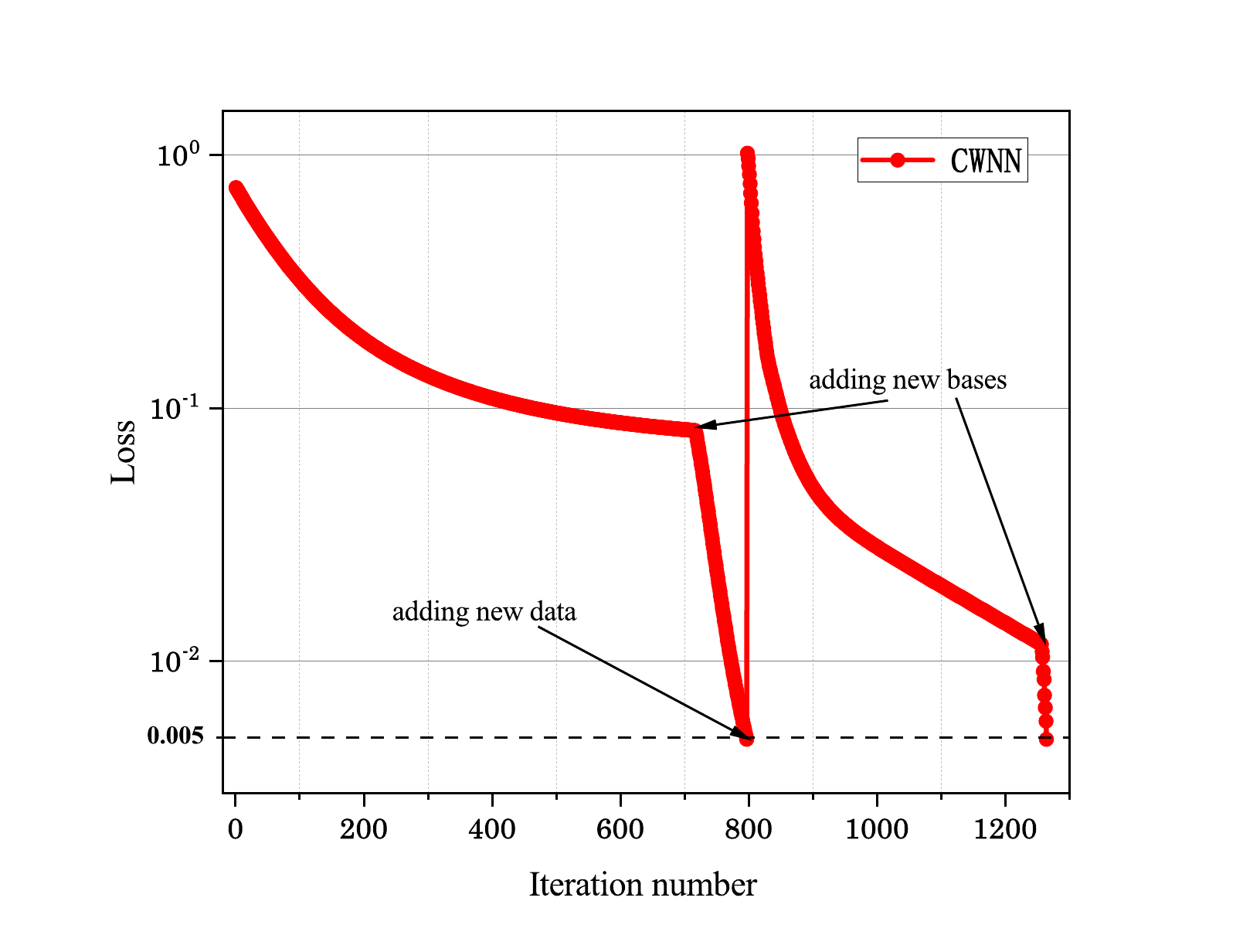}
\caption{The CWNN is first trained using dataset $DS_1$ until the $798^{\text{th}}$ iteration, when the loss satisfies $Loss \leq 0.005$. After the $798^{\text{th}}$ iteration, dataset $DS_2$ is incorporated. The CWNN is then retrained to accommodate the newly introduced data set until the updated network again reaches the desired accuracy $\varepsilon = 0.005$}.
\label{fignew9}
\end{figure}
As illustrated in Fig.\,\ref{fignew9}, the network is first trained with dataset $DS_1$ until the loss reaches the desired accuracy threshold, $Loss \leq 0.005$. When dataset $DS_2$ is subsequently incorporated, a sharp increase in the loss, as defined in \eqref{lost_function}, is observed. At the $1255^{\text{th}}$ iteration, the addition of new bases enables the loss to reconverge to 0.005. This behavior highlights CWNN's capability to approximate nonlinear functions effectively using combined datasets and adaptive wavelet bases.

\noindent {\underline{Example 3}}: On-line Learning from Time-series Data

Neural networks offer efficient approximation for nonlinear time series. For example, \cite{9431777} proposed a nonlinear autoregressive model with $y_t$ depending on past values, which can be approximated by a WNN. Unlike Examples 1 and 2, data here is collected online. Both WNN and CWNN can learn from online measurements, but CWNN achieves higher computational efficiency. To demonstrate this, the nonlinear function $f(\cdot)$ is modified during training, and CWNN adapts by incrementally adding wavelet bases to maintain accuracy.

We consider the following two-dimensional mapping:
$y= \arctan(\pi(x_1^2 + x_2^2))^{\frac{1}{2}}$.
Then, we obtain the following nonlinear auto-regression:
\begin{align}
\label{eq24}
    y_t = \arctan(\pi(y_{t-1}^2 + y_{t-2}^2))^{\frac{1}{2}} + d_3(t),
\end{align}
where $d_3(t)$ represents the measurement noises. They follow a normal distribution with mean zero and standard deviation to be $0.01$. By setting $y_1 = y_2 = 1$, we generate the time series ${y_t}$ using \eqref{eq24} and approximate the nonlinear mapping with both WNN and CWNN methods.

For online training at time $t$, we use $y_{t-1}$ and $y_{t-2}$ as inputs, with $y_t$ as the actual output. We train the WNN and CWNN using a window of 10 sampling instants. The time step is 1 second. After every 10 seconds, both networks adjust their structure and weights using 10 collected input/output pair. This process repeats as more data is collected. In the CWNN, we set $\kappa = 0.36$ and $\iota_r = 0.0001$. Translation centers $\{{\boldsymbol{K}}_{1,1},{\boldsymbol{K}}_{1,2},\dots,{\boldsymbol{K}}_{{1,N_1}}\}$ for Algorithm \ref{alg1} are given as $\{ 0, 0.5, 1, 1.5, 2 \} \times \{ 0, 0.5, 1, 1.5, 2 \}$, with initial resolution $m_{init} = 2$ and wavelet frequency $2^2$. These parameters are passed to Algorithm \ref{alg2} with translation centers ${\boldsymbol{K}}^{\boldsymbol{m_{init}}}_{in}$ are $\{ 0, 0.25, 0.5, \dots, 2 \} \times \{ 0, 0.25, 0.5, \dots, 2 \}$. We select $\mu = 1/3$, $\varepsilon = 0.02$, and termination thresholds as $0.00004$. We train \eqref{eq24} online using time series data ${y_t}$, achieving desired accuracy with wavelet bases at frequency $2^2$.

\begin{figure}[!t]
\centering
\subfigure[]{\includegraphics[width=0.35\textwidth]{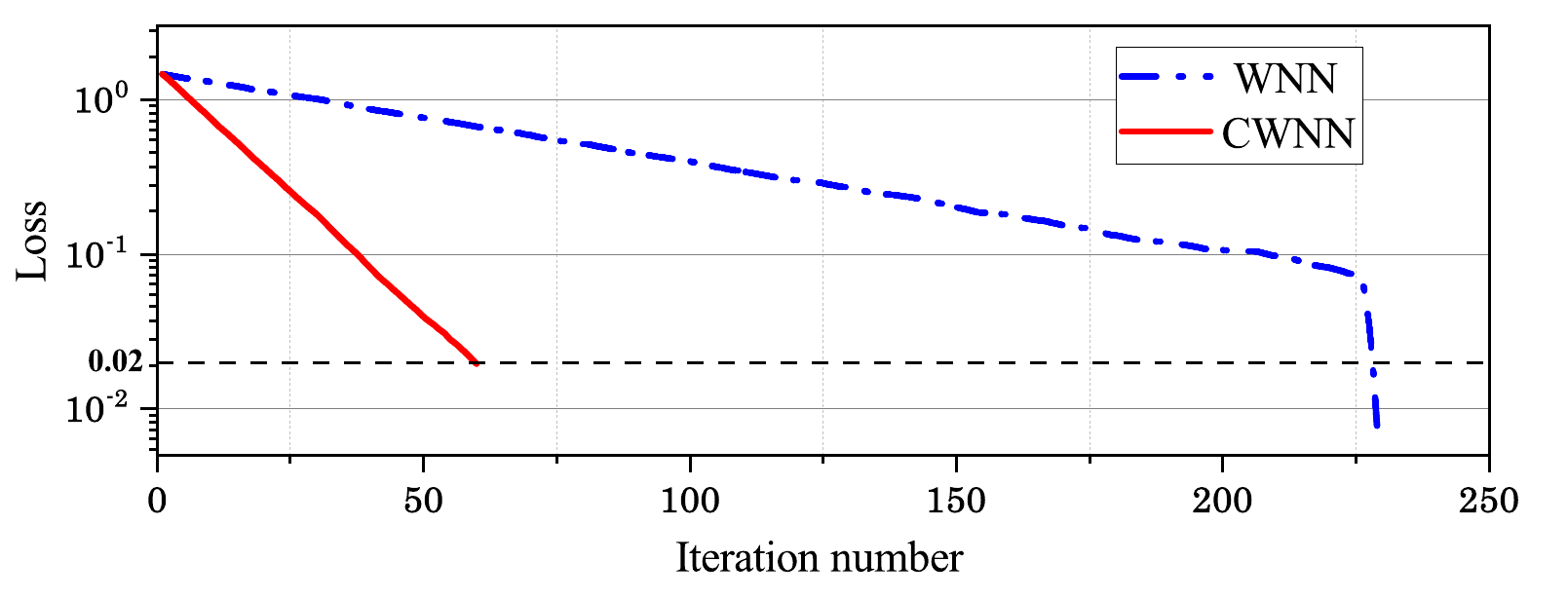}
\label{fignew5a}}
\subfigure[]{\includegraphics[width=0.40\textwidth]{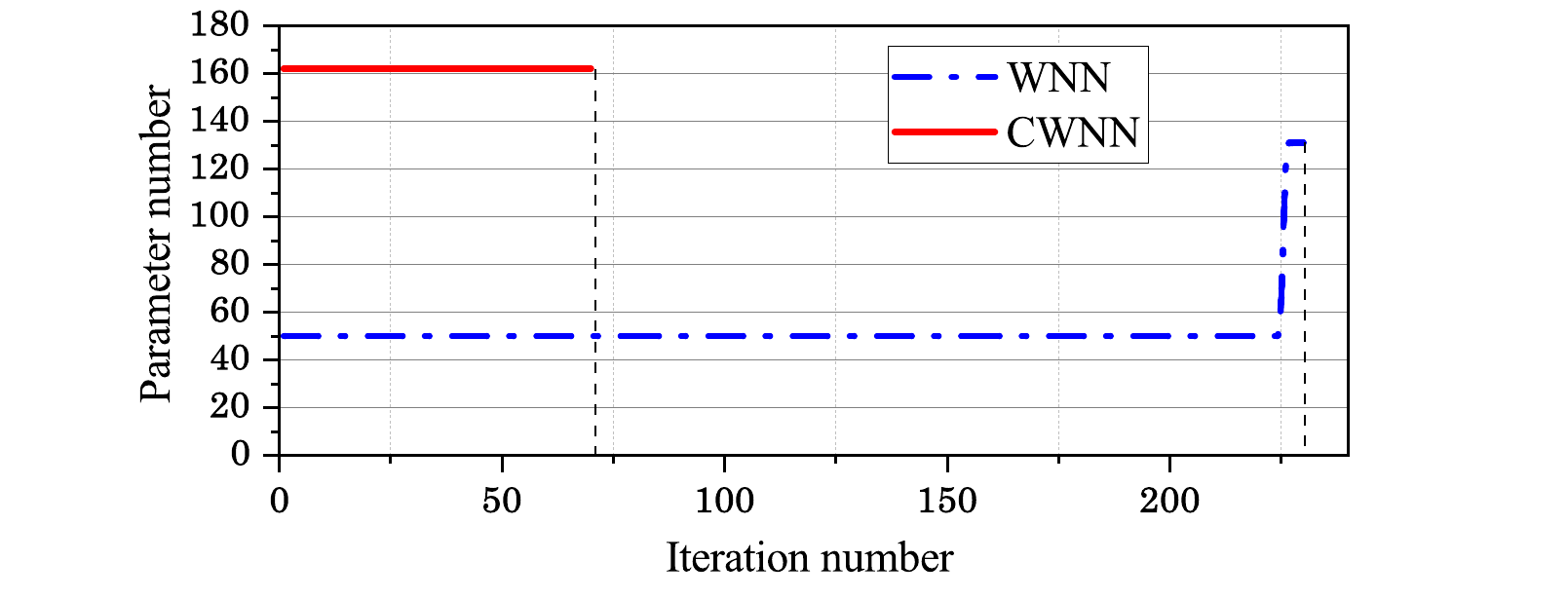}
\label{fignew5b}}
\caption{(a) Loss\eqref{eq24}. (b)  the parameter number used by WNN and CWNN until they reach the predefined accuracy $\varepsilon=0.02$.}
\label{fignew5}
\end{figure}

In Fig.\,\ref{fignew5}, we compare the Loss and wavelet number for WNN and CWNN. The iteration number corresponds to the data collection cycles (e.g., two iterations mean 20 input-output data pairs have been collected). Fig.\,\ref{fignew5a} shows that CWNN's Loss converges faster than WNN's. Fig.\,\ref{fignew5b} indicates CWNN has lower computational requirements during training process. The final parameter counts for both models are comparable.

At $t=10$, the nonlinear mapping \eqref{eq24} is modified to
\bea
    \label{eq25}
    y_t &=& \arctan(\pi(y_{t-1}^2 + y_{t-2}^2))^{\frac{1}{2}}
    \nonumber\\
    & &+ \cos(\pi(y_{t-1}^2 + y_{t-2}^2)) +  d_3(t),
\eea
where the measurement noises $d_3(t)$ are the same as those from (\ref{eq24}).

\begin{figure}[!t]
\centering
\includegraphics[width=0.4\textwidth]{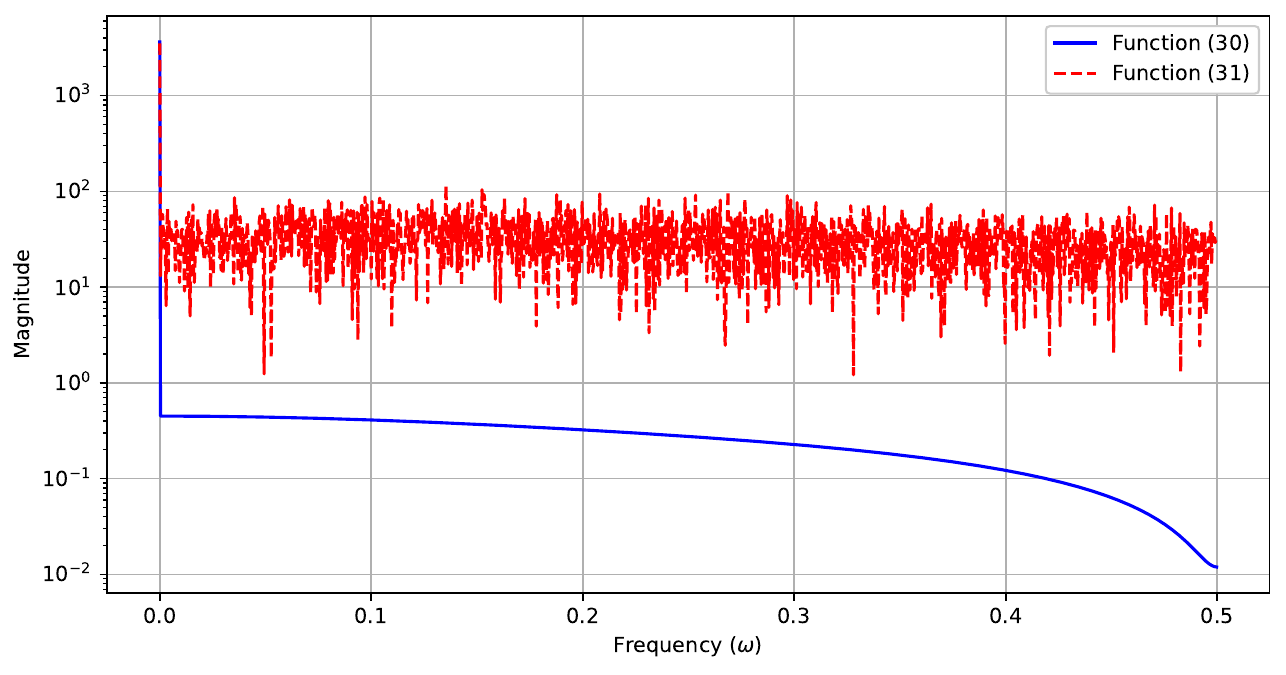}
\caption{the spectra of functions \eqref{eq24} and \eqref{eq25}.}
\label{image_spectrum_comparison}
\end{figure}

To highlight the spectral differences between functions \eqref{eq24} and \eqref{eq25}, their spatial frequency spectra are shown in Fig.~\ref{image_spectrum_comparison}. The spectrum of the nonlinear function in \eqref{eq24} is concentrated near $\omega=0$ and rapidly decays by $\omega=0.5$, whereas the nonlinear function in \eqref{eq25} maintains energy across the entire frequency range. This indicates a clear frequency shift when the system changes from \eqref{eq24} to \eqref{eq25} at $t=10$.  

\begin{figure}[!htb]
\centering
\includegraphics[width=0.45\textwidth]{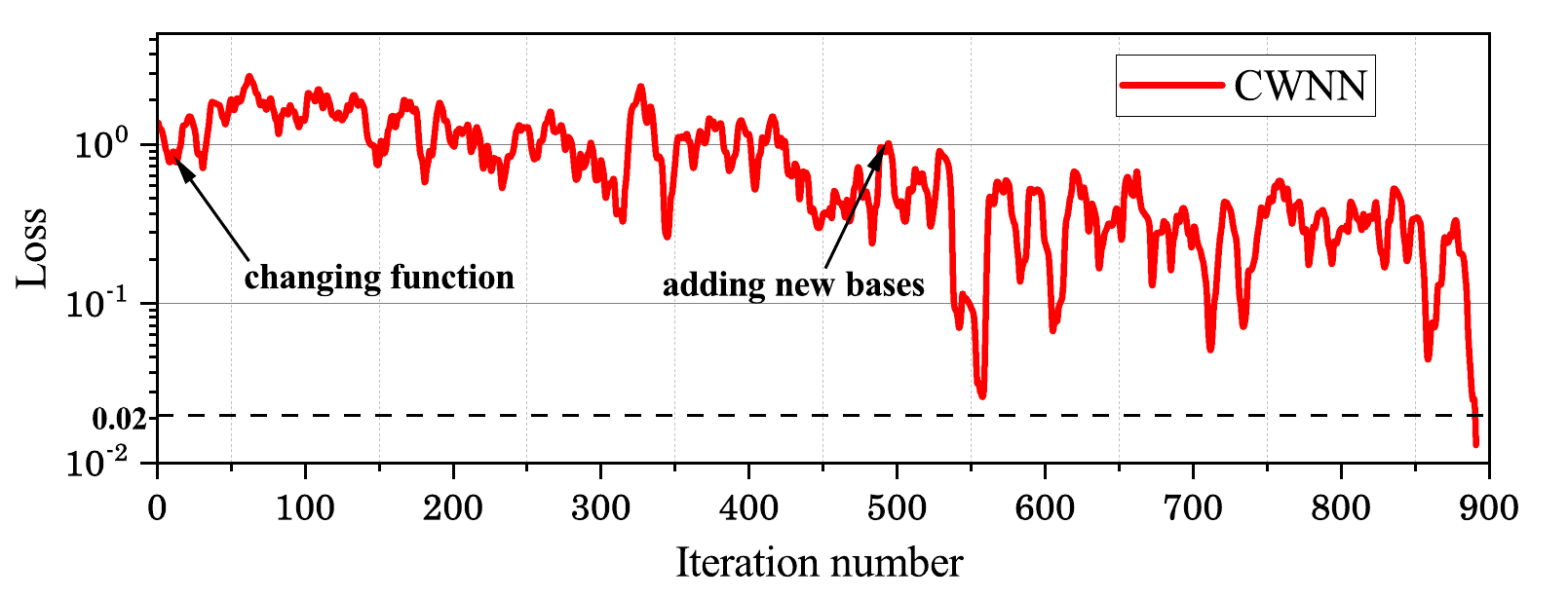}
\caption{The computed Loss  \eqref{eq24} (before $10^{th}$ iteration and for \eqref{eq25} after $10^{th}$ iteration}
\label{fignew6}
\end{figure}

Fig.~\ref{fignew6} shows the loss evolution as the nonlinear mapping changes. The mapping is modified at the $8^{\text{th}}$ iteration, and new wavelets are added at the $493^{\text{rd}}$ iteration. Despite the change, the loss converges to the target accuracy by the $891^{\text{st}}$ iteration, demonstrating the framework’s adaptability to time-varying mappings. Since the difference between the two functions is small, only a modest adjustment—adding new wavelets at the $493^{\text{rd}}$ iteration—is observed.

\noindent \underline{Example 4}: Learning a high-dimensional mapping from a real-world time-series dataset (publicly available)

Instead of simulated data as in Examples 1–3, a real-world time-series dataset \cite{Agogino_data_set_2007} from the BEST Lab at UC Berkeley is used to evaluate the practicality of the proposed CWNN. The dataset captures experiments on a metal milling machine under varying operating conditions, including different spindle speeds, feed rates, and depths of cut. It records tool wear progression through flank wear measurements ($VB$) and is widely used in research on tool wear prediction, anomaly detection, and predictive maintenance. This dataset has served as a benchmark for evaluating machine learning algorithms in manufacturing \cite{Lu_Lei2023}.

In this study, we utilize the Milling Dataset with Sensor and Condition Information (MDSCI), following the approach in~\cite{Lu_Lei2023}. The MDSCI contains 167 experimental trials conducted under 16 distinct machining configurations, resulting in over 160{,}000 temporally ordered observations.  Each trial includes synchronized sensor measurements sampled at 250~Hz: AC spindle motor current (\( \mathit{smcAC} \)), DC spindle motor current (\( \mathit{smcDC} \)), table vibration (\( VT \)), and spindle vibration (\( VS \)). Additionally, flank wear values (\( VB \)) are manually annotated at irregular intervals throughout the trials. For the purpose of temporal degradation modeling, \( VB \) is treated as the target variable. As input features, we compute the standard deviation and kurtosis of each of the four sensor signals (\( \mathit{smcAC} \), \( \mathit{smcDC} \), \( VT \), and \( VS \)), yielding a total of eight input features.

An unknown nonlinear function characterizes the relationship between the current lank wear (\( VB_t \)) and 9 input (including 8 current input features and the past  flank wear (\( VB_{t-1} \)).
\begin{align}
\label{syst_4}
\mathrm{VB}_t = f\left(\mathrm{VB}_{t-1}, \boldsymbol{x}_t \right),   
\end{align}
where the input vector $\boldsymbol{x}_t$ denote the eight current process input features.

Min-Max scaling \cite{Lu_Lei2023} is used to normalize the input features $\boldsymbol{x}_t$. $80\%$ of the data is randomly assigned for training, with the remaining $20\%$ used for testing. Notably, the input dimension of mapping \eqref{syst_4} is nine and the output dimension is one. Three GNN-based methods (GNN-3, GNN-5, GNN-7), along with WNN and CWNN, are evaluated for comparison. The parameter configurations and network architectures of the GNN-based models are consistent with those in Example 1.

In the CWNN, parameters $\kappa$, $\iota_r$, and $\varepsilon$ are set to $\frac{2}{3}$, $0.001$, and $0.015$, respectively. First, calculate the initial frequency using Algorithm \ref{alg1}.
Translation centers $\{{\boldsymbol{K}}_{1,1},{\boldsymbol{K}}_{1,2},\dots,{\boldsymbol{K}}_{{1,N_1}}\}$ are given as $\{(s_1,s_2, \dots,s_9 ) \vert s_i \in \{ 0,0.5,1 \}, i = 1,2 \dots, 9\}$.  Since $\kappa=\frac{2}{3}$, $\{{\boldsymbol{K}}_{1,1},{\boldsymbol{K}}_{1,2},\dots,{\boldsymbol{K}}_{{1,N}_{J}}\} =
\{(s_1,s_2, \dots,s_9 ) \vert s_i \in \{ 0,1 \}, i = 1,2 \dots, 9\}$. 
From Algorithm \ref{alg1}, the initial resolution $m_{init}=0$ and initial frequency $2^0$ are obtained. The initial frequency $2^0$ is input into Algorithm \ref{alg2}, with translation centers 
${\boldsymbol{K}}^{\boldsymbol{m_{init}}}_{in} = 
\{(s_1,s_2, \dots,s_9 ) \vert s_i \in \{ 0,1 \}, i = 1,2 \dots, 9\}$. 
Termination thresholds are all set to $0.00004$, and $\mu=\frac{1}{3}$. Finally, we apply Algorithm \ref{alg2}.

\begin{figure}[htb!]
\centering
\subfigure[]{\includegraphics[width=0.30\textwidth]{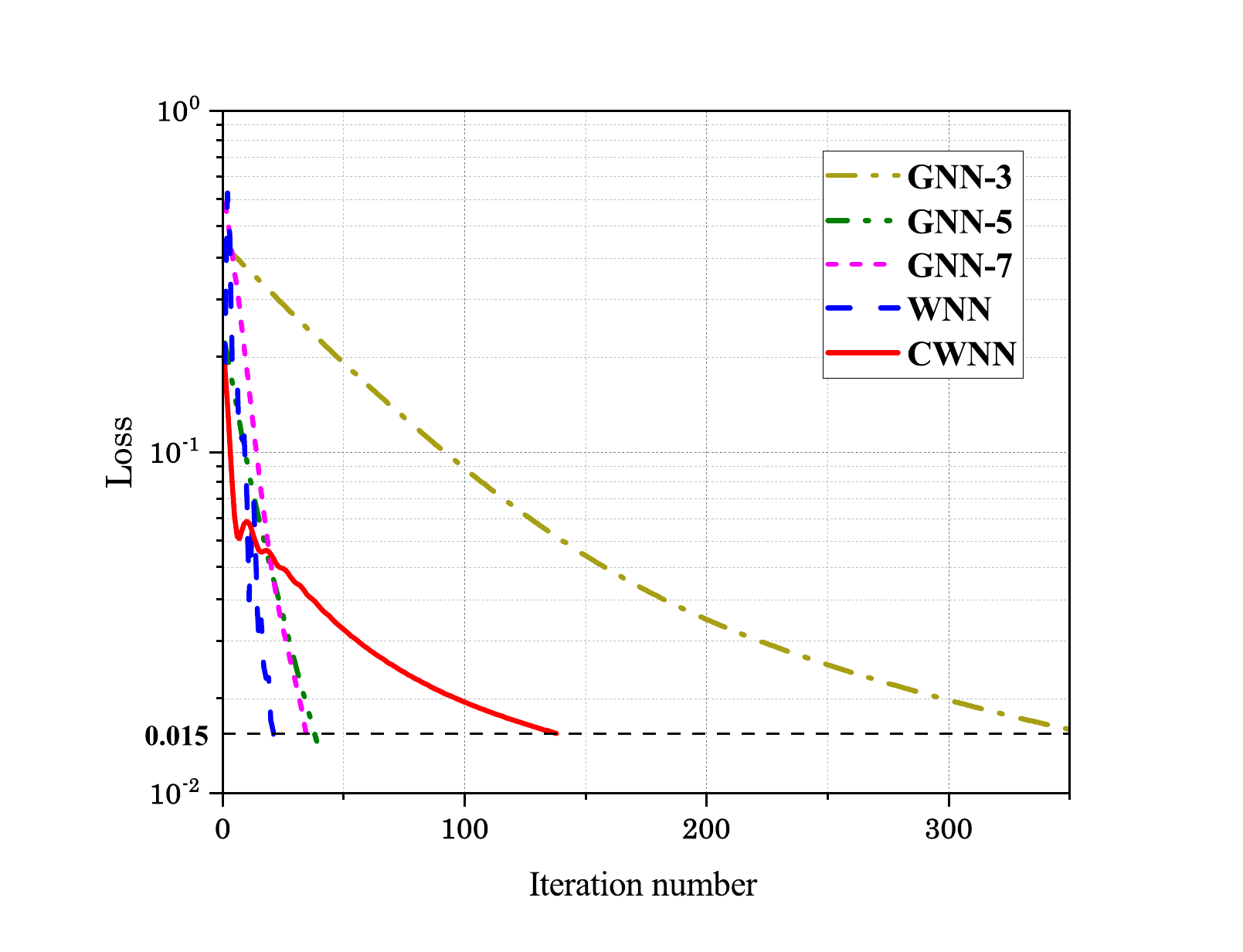}
\label{image_highdim_loss}}
\subfigure[]{\includegraphics[width=0.32\textwidth]{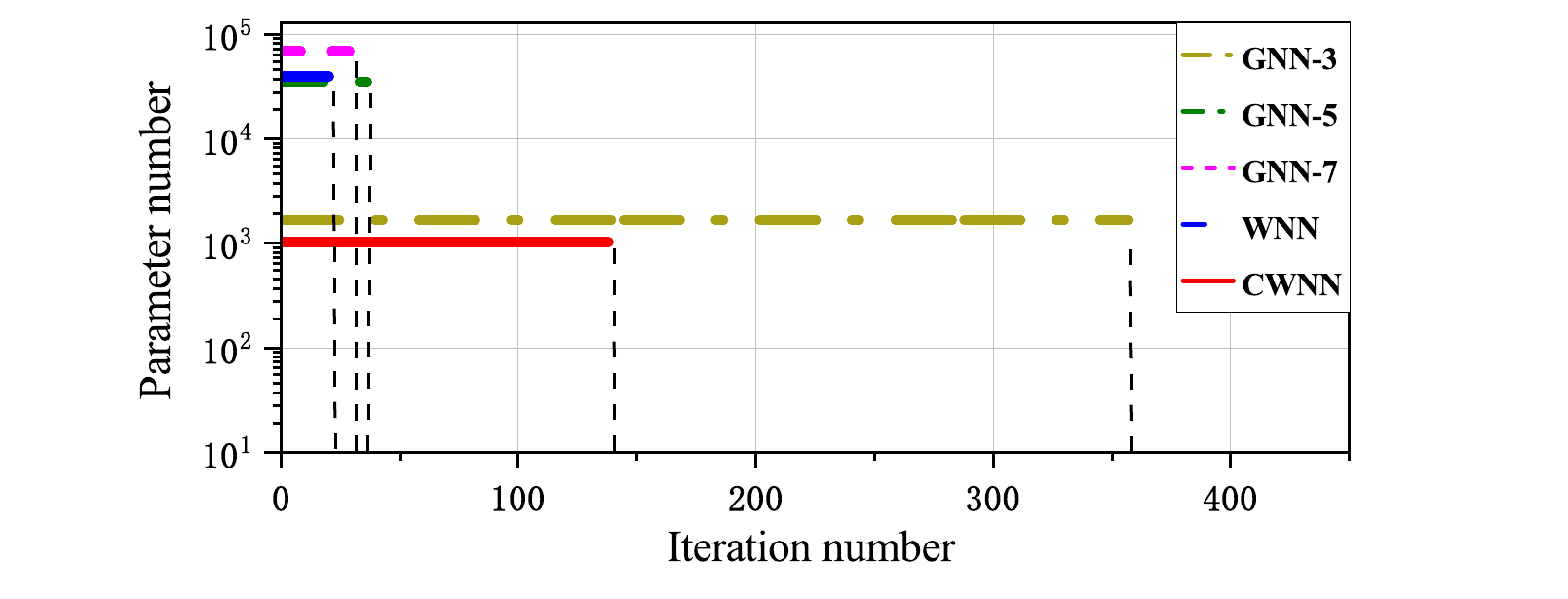}
\label{image_highdim_paras}}
\caption{ (a) The computed Loss using the training set to learn unknown mapping in \eqref{syst_4}. (b)  The number of parameters used by 5 NNs.}
\label{fig_highdim}
\end{figure}

Fig.\,\ref{fig_highdim} shows the computed Loss (\ref{lost_function}) and number of parameters used by five NNs during training. The three GNN-based methods (GNN-3, GNN-5, GNN-7) reach the accuracy threshold \(\varepsilon = 0.015\) at the 357\textsuperscript{th}, 39\textsuperscript{th}, and 35\textsuperscript{th} epochs, requiring 1,665, 35,201, and 68,737 trainable parameters, respectively—each far exceeding the 1,510 parameters needed by CWNN. Both WNN and CWNN also achieve the threshold, requiring 21 and 138 epochs, respectively. However, WNN demands 39,366 parameters, over 30 times more than CWNN.

We also test the five trained NNs on the testing set, with results summarized in Table \ref{tablenew4}. Again, GNN-7 performs the worst, indicating overfitting, while the other NNs demonstrate a reasonable selection of parameters.

\begin{table}[!htb]
\renewcommand{\arraystretch}{1.5}
\caption{Accuracy with testing datasets }
\label{tablenew4}
\centering
\setlength{\tabcolsep}{3pt}
\begin{tabular}{p{50pt}|p{30pt}|p{30pt}|p{30pt}|p{30pt}|p{30pt}}
\hline
\textbf{Method} &
\textbf{ GNN-3} &
\textbf{ GNN-5} &
\textbf{ GNN-7} &
\textbf{ WNN} &
\textbf{ CWNN} \\
\hline 
\textbf{Accuracy} & $0.0341$ & $0.0363$ & $0.2673$ & $\bf 0.0142$ & $0.0213$\\
\hline
\end{tabular}
\end{table}

These results underscore CWNN's superior computational efficiency and effectiveness, achieving a favorable balance between fast convergence and minimal parameter overhead.

\section{Discussion}\label{discussion}

Our work introduces a novel constructive wavelet neural network framework with two distinct advantages: an efficient method for identifying the initial basis function and a unique mechanism for expanding the wavelet basis set. We demonstrate its effectiveness and broad applicability through five examples. These examples, coupled with our theoretical analysis, highlight the flexibility of the proposed CWNN, demonstrating its potential in practical machine learning applications. Future research will explore deploying CWNNs in real-world applications, such as medical diagnosis \cite{lu2024decoding}, mechanical fault detection \cite{lu2022weak}, and multi-agent control \cite{zhou2022iterative}.

We acknowledge that designing neural networks typically involves a trial-and-error process, which is time-consuming even for experienced researchers in searching for an optimal architecture. Many existing neural architecture search (NAS) methods rely on heuristic algorithms, such as reinforcement learning (RL) \cite{gao2022efficient}, and AutoNet-generated networks \cite{shen2024autonet}, to guide the architecture search process. These methods typically prioritize monitoring and minimizing the validation loss of the cost function by adding more neural network layers, leading to network designs primarily focused on increasing layers. In contrast, our proposed CWNN framework introduces a novel mechanism for searching for the wavelet basis set. It enables automatic adjustments of wavelet bases based on dataset complexity, facilitating the exploration of more efficient neural networks. Our future work will delve into evaluating the performance of our developed CWNN framework in seeking optimal deep neural networks and conducting comparative studies with these state-of-the-art NAS approaches such as \cite{gao2022efficient,shen2024autonet}.

\section{Conclusions}\label{sec6}

This work proposed a wavelet neural network (WNN) framework for approximating nonlinear functions with pre-defined accuracy by estimating the frequency information of this unknown mapping. The framework starts by estimating the initial wavelet frequency and proves that approximation error is bounded by the function's energy and the number of wavelet bases. Our wavelet-basis increase algorithm locates the time-frequency concentration domain of the function and adds necessary bases.  The main result established a relationship between approximation error and function energy in $L^2(\textbf{R}^d)$ space, optimizing approximation performance.
Five simulation examples show the proposed algorithms can improve the convergence speed and reduce the computational complexity for learning a large class of complex nonlinear functions. Future work will explore incorporating adaptive WNNs to learn the weights of WNNs for a more general class of dynamic systems.

\section{Appendix}\label{sec7}

\subsection{Proof of Proposition \ref{proposition1}}\label{appendixA}
\emph{Proof:} We first calculate the coefficients ${C}_{mn}$ assuming that $f(\cdot)$ is known. To this end, we consider an arbitrary wavelet base, denoted by ${\psi }_{mn}^{\ast}(\boldsymbol{x})$, from the set $\mathcal{W}$ defined in \eqref{eq2}.
\begin{align}
\label{eq2}
\mathcal{W}\triangleq\left\{\psi_{mn}(\boldsymbol{x})=2^{\frac{dm}{2}}\psi(2^{m}\boldsymbol{x}-n)|m \in \textbf{Z}, n \in \textbf{Z}^d,\boldsymbol{x}\in \textbf{R}^d\right\},
\end{align}
Then, we have
\[\left\langle f(\boldsymbol{x}),{\psi }_{mn}^{\ast}(\boldsymbol{x})\right\rangle=\sum_m\sum_n{C_{mn}\left\langle{\psi}_{mn}(\boldsymbol{x}),{\psi }_{mn}^{\ast}(\boldsymbol{x}) \right\rangle},\]
As the wavelet bases are orthogonal, the coefficient of ${\psi }_{mn}^{\ast}(\boldsymbol{x})$ in the expansion of $f(\boldsymbol{x})$, denoted as ${C}_{mn}^{\ast}$, are obtained as follows:
\begin{align*}
C_{mn}^{\ast}=\frac{\left\langle f(\boldsymbol{x}),{\psi }_{mn}^{\ast}(\boldsymbol{x})\right\rangle}{\left\langle {\psi}_{mn}^{\ast}(\boldsymbol{x}),{\psi }_{mn}^{\ast}(\boldsymbol{x})\right\rangle}=\frac{\int_{\textbf{R}^d}{f(\boldsymbol{x})\cdot{\psi }_{mn}^{\ast}(\boldsymbol{x})\ d{\boldsymbol{x}}}}{\int_{\textbf{R}^d}{{\psi }_{mn}^{\ast}(\boldsymbol{x})\cdot{\psi }_{mn}^{\ast}(\boldsymbol{x})\ d{\boldsymbol{x}}}}.
\end{align*}

Therefore, $f(\boldsymbol{x})$ can be expressed as
\begin{align}
f(\boldsymbol{x})=\sum_m\sum_n{\frac{\int_{\textbf{R}^d}{f(\boldsymbol{x})\cdot{\psi }_{mn}(\boldsymbol{x})\ d{\boldsymbol{x}}}}{\int_{\textbf{R}^d}{{\psi }_{mn}(\boldsymbol{x})\cdot{\psi }_{mn}(\boldsymbol{x})\ d{\boldsymbol{x}}}}}{\psi }_{mn}(\boldsymbol{x}), \nonumber
\end{align}
and accordingly, for any $m$, ${f}_m$ can be expressed as
\[{f}_m(\boldsymbol{x})=\sum_n{{C_{mn}\psi }_{mn}}(\boldsymbol{x}).\]

The energy of ${f}_m$ is calculated as
\begin{align}
E_m=&\int_{\textbf{R}^d}{f^2_m(\boldsymbol{x})d{\boldsymbol{x}}}\nonumber\\
=&\int_{\textbf{R}^d}{{\left(\sum_n{{C_{mn}\psi }_{mn}(\boldsymbol{x})}\right)}^2d{\boldsymbol{x}}}\nonumber\\
 =&\int_{\textbf{R}^d}{\sum_n{C^2_{mn}{\psi }^2_{mn}(\boldsymbol{x})}d{\boldsymbol{x}}}\nonumber\\
 =&\sum_n{\frac{{(\int_{\textbf{R}^d}{f_m(\boldsymbol{x})\cdot{\psi}_{mn}(\boldsymbol{x})\ d{\boldsymbol{x}}})}^2}{\int_{\textbf{R}^d}{{\psi}^2_{mn}(\boldsymbol{x})\ d{\boldsymbol{x}}}}}.\nonumber
\end{align}

Note that $\int_{\textbf{R}^d}{{f}_m(\boldsymbol{x})\cdot{\psi }_{mn}(\boldsymbol{x})d{\boldsymbol{x}}}=C_{mn}{\|{\psi }_{mn}(\boldsymbol{x})\|^2}$. The energy $E_m$ can then be simplified as
\[E_m=\sum_n{{C^2_{mn}\|{\psi }_{mn}(\boldsymbol{x})\|}^2},\]
where $C_{mn}$ represent the wavelet-basis coefficients in ${W}_m$ space. The proof is completed. $\hfill\square$

\subsection{Proof of Theorem \ref{thm1}}\label{appendixB}

Before proving Theorem \ref{thm1}, we present a technical lemma.
Consider a function $\widehat{\phi}$ satisfying \eqref{eq3}, \eqref{eq4} and \eqref{eq5}. 
\begin{align}
\label{eq3}
   & \widehat{\psi}(\boldsymbol{\omega})=\widehat{\phi}\left(\abs{\boldsymbol{\omega}}\right),\\
\label{eq4}
    &\inf_{|\omega|\in[1,2]}\sum_{m\in \textbf{Z}}\left|\widehat{\phi}\left(2^{-m}\omega\right)\right|^2\ > 0\\
\label{eq5}
    &\left|\widehat{\phi}(\omega)\right| \leq C|\omega|^{\alpha}\left(1+|\omega|^2\right)^{-\frac{\gamma}{2}},
\end{align}
For $m\ge m_0\ (m_0\ge 1)$, define ${\beta }_1(\eta )$ as
\[
{\beta }_1(\eta )\triangleq \mathop{sup}_{{ | \boldsymbol{\omega} | } \in [1,2]}\sum_{m\ge m_0}\left|\widehat{\phi }\left({| 2^{-m}\boldsymbol{\omega} |} \right)\right|\cdot \left|\widehat{\phi }\left({| 2^{-m}\boldsymbol{\omega} +\eta |}\right)\right| ,
\]
and for $m{\le m}_1\ (m_1\le 0)$, define ${\beta }_2(\eta )$ as
\[
{\beta }_2(\eta )\triangleq \mathop{sup}_{{| \boldsymbol{\omega} | }\in [1,2]}\sum_{m\le m_1}\left|\widehat{\phi }\left({|2^{-m}\boldsymbol{\omega} |}\right)\right|\cdot \left|\widehat{\phi }\left({|2^{-m}\boldsymbol{\omega} +\eta |} \right)\right|,
\]
where $\eta \in \textbf{Z}^d$, then we have Lemma \ref{lemma2}.
\begin{lemma}[\cite{26}, Appendix-C]\label{lemma2}
\begin{align}
\label{eq16}
{\beta }_1(\eta )
\le& C_1{\left(1+{|\eta |}^2\right)}^{-\frac{\gamma -\alpha }{2}}\displaystyle\sum_{m=m_0}^{{\infty}}{\ }2^{-m\alpha},\\
\label{eq17}
{\beta }_2(\eta )
\le& C_2{\left(1+{|\eta |}^2\right)}^{-\frac{d\left(1+{\varepsilon }_2\right)}{2}}\sum_{m=-\infty}^{m_1}{\left(2^{-\frac{\delta \left(\gamma -\alpha \right)}{2}}\right)}^{-m},
\end{align}
where $C_1>0,C_2>0$, ${\varepsilon }_2$$>0$, $\gamma >\alpha +d$, $\alpha >0$.
\end{lemma}

\emph{Proof:} We prove Theorem \ref{thm1} in three steps. First, we disassemble \eqref{eq15} into four terms and prove that the first and second terms depend on the energy of $f$, and the third and fourth terms depend on wavelets with frequencies and translation centers outside $B_\varepsilon$, respectively. Second, we prove that for a given approximation accuracy $\varepsilon$, there exists some $B_\varepsilon$ in which the third term can be bounded by $\varepsilon$. Finally, we prove that the fourth term is similarly bounded by $\varepsilon$.

First, we introduce the Plancherel theorem,
\begin{align}
\langle f,g\rangle =\int_{{\mathrm{R}}^d}\overline{f({\boldsymbol{x}})}g(\boldsymbol{x})d\boldsymbol{x}
 =\int_{{\mathrm{R}}^d}\overline{\widehat{f}(\boldsymbol{\omega} )}\widehat{g}(\boldsymbol{\omega} )d\boldsymbol{\omega} =\langle \widehat{f},\widehat{g}\rangle, \nonumber
\end{align}
where $f, g \in L^2(\textbf{R}^d)$, $\widehat{f}$ and $\widehat{g}$ are the Fourier transform of $f$ and $g$. Combined with the Plancherel theorem, we have
\begin{align}
&\left\|f-\sum_{\left(m,n\right)\in B_{\varepsilon }}{\left\langle {\psi }_{mn},f\right\rangle {\psi }_{mn}}\right\| \nonumber\\
&=\mathop{\mathrm{sup}}_{\|f_{\mathrm{1}}\|\mathrm{=1}}\left|\sum_{\left(m,n\right)\notin B_{\varepsilon }}{\left\langle f,{\psi }_{mn}\right\rangle \left\langle f_{\mathrm{1}},{\psi }_{mn}\right\rangle }\right|\nonumber\\
&\mathrm{\le }\mathop{\mathrm{sup}}_{\|f_{\mathrm{1}}\|\mathrm{=1}}\sum_{{{ \begin{array}{c}
n\mathrm{\in }{\textbf{Z}^d},m\mathrm{\in }{\mathit{\Lambda}}_{re} \end{array}
}}}\left|\left\langle f_{\mathrm{1}},{\psi }_{mn}\right\rangle \right|\nonumber\\
&\textrm{·}\left[\left|\left\langle {\widehat{\psi} }_{mn},\left({\mathcal{P}}_{{\boldsymbol{\Omega }}_{1}}-\mathcal{P}_{{\boldsymbol{\Omega }}_0}\right)\widehat{f}\right\rangle \right|+\left|\left\langle {\widehat{\psi} }_{mn},\left(\boldsymbol{1}-{\mathcal{P}}_{{\boldsymbol{\Omega }}_1} + {\mathcal{P}}_{{\boldsymbol{\Omega }}_0}\right)\widehat{f}\right\rangle \right|\right]\nonumber\\
&+\mathop{\mathrm{sup}}_{\|f_{1}\|=1}\sum_{{{ \begin{array}{c}
n \in {\mathit{\Lambda}}_{tr},
m \in \textbf{Z} \end{array}
}}}\left|\left\langle f_{\mathrm{1}},{\psi }_{mn}\right\rangle \right|\ \nonumber\\
&\textrm{·}\left[\left|\left\langle {\psi }_{mn},\left({\mathcal{Q}}_{\boldsymbol{T}}\right)f\right\rangle \right|\mathrm{+}\left|\left\langle {\psi }_{mn},\left(\boldsymbol{1}-\mathcal{Q}_{\boldsymbol{T}}\right)f\right\rangle \right|\right],\nonumber
\end{align}
where ${\mathit{\Lambda}}_{re}=\left\{m\mathrel{\left|\vphantom{m\ m\ge m_0}\right.\kern-\nulldelimiterspace}m\ge m_0,m\in \textbf{Z}\right\}\cap \left\{m\right|m\le m_1,m\in \textbf{Z}\}$, and ${\mathrm{\Lambda }}_{tr}=\left\{n\right|\textbf{Z}^d\backslash (|n|_s \preceq 2^{m} \boldsymbol{T} + \boldsymbol{t}_{\varepsilon})\}$ represents the $d$-dimensional integer set $\textbf{Z}^d$ excluding $|n|_s \preceq 2^{m}\boldsymbol{T} + \boldsymbol{t}_{\varepsilon}$, $\widehat{\psi}_{mn}$ is the Fourier transform of ${\psi}_{mn}$. As the wavelets are orthonormal, we obtain
\begin{align}
\label{eq18}
&\left\|f-\sum_{\left(mn\right)\in B_{\varepsilon }}{\left\langle {\psi }_{mn},f\right\rangle {\psi }_{mn}}\right\|\nonumber\\
&\le \|\left(\boldsymbol{1}-{\mathcal{P}}_{{\boldsymbol{\Omega}}_{\mathrm{1}}}\mathrm{+}{\mathcal{P}}_{{\boldsymbol{\Omega}}_0}\right)\widehat{f}\|\mathrm{+}\|\left(\boldsymbol{1}-{\mathcal{Q}}_{\boldsymbol{T}}\right)f\|\nonumber \\
&+{\left[\sum_{{{ \begin{array}{c}
n\mathrm{\in }\textbf{Z}^d,m\mathrm{\in }{\mathit{\Lambda}}_{re} \end{array}
}}}{\left|\left\langle {\widehat{\psi} }_{mn},\left({\mathcal{P}}_{{\boldsymbol{\Omega}}_{\mathrm{1}}}\mathrm{-}{\mathcal{P}}_{{\boldsymbol{\Omega}}_0}\right)\widehat{f}\right\rangle \right|}^{\mathrm{2}}\right]}^{\frac{\mathrm{1}}{\mathrm{2}}} \nonumber\\
&+{\left[\sum_{{{{{ \begin{array}{c}
n\mathrm{\in }{\mathit{\Lambda}}_{tr},m\mathrm{\in }\textbf{Z} \end{array}
}}}}}{\left|\left\langle {\psi }_{mn},{\mathcal{Q}}_{\boldsymbol{T}}f\right\rangle \right|}^{\mathrm{2}}\right]}^{\frac{\mathrm{1}}{\mathrm{2}}}.
\end{align}

First, consider the third term on the right-hand side of Eq. \eqref{eq18}. Combining the Poisson formula
\[\sum_{n\mathrm{\in }{\textbf{Z}^d}}{\mathrm{\ }}e^{icn^{\mathrm{T}}{\boldsymbol{x}}}\mathrm{=}\left(\frac{\mathrm{2}\pi }{c}\right)\prod_{j\mathrm{=1}}^d{\mathrm{\ }}\sum_{n_j\mathrm{\in }\textbf{Z}}\delta \left(x_j\mathrm{-}\frac{\mathrm{2}\pi }{c}n_j\right),\]
we get
\begin{align}
&\sum_{{{ \begin{array}{c}
n\mathrm{\in }{\textbf{Z}^d},m\mathrm{\in }{\mathit{\Lambda}}_{re} \end{array}
}}}{\left|\left\langle {\widehat{\psi} }_{mn},\left({\mathcal{P}}_{{\boldsymbol{\Omega}}_{1}}-{\mathcal{P}}_{{\boldsymbol{\Omega}}_0}\right)\widehat{f}\right\rangle \right|}^{2}\nonumber\\
&={\left({2\pi }\right)}^d\sum_{{{ \begin{array}{c}
l\mathrm{\in }{\textbf{Z}^d},m\mathrm{\in }{\mathit{\Lambda}}_{re} \end{array}
}}}{\int_{ \begin{array}{c}
{\boldsymbol{\omega}}\mathrm{\in }{\mathit{\Lambda}}_{fr}\end{array}
}}\overline{\widehat{\psi }\left(2^{-m}\boldsymbol{\omega} \right)} \nonumber \\
&\cdot \widehat{\psi }(2^{-m}\boldsymbol{\omega} -{2\pi }l) \overline{\widehat{f}(\boldsymbol{\omega} )}\widehat{f}(\boldsymbol{\omega} -\frac{\mathrm{2}\pi }{2^{-m}}l)d\boldsymbol{\omega}, \nonumber
\end{align}
where ${\mathrm{\Lambda }}_{fr}=\{\boldsymbol{\omega} \ |{\boldsymbol{\Omega}}_0\le \left|\boldsymbol{\omega} \right|\le {\boldsymbol{\Omega}}_1\}\cap \{\boldsymbol{\omega} \ |{\boldsymbol{\Omega}}_0\le \left|\boldsymbol{\omega} -\frac{\mathrm{2}\pi }{2^{-m}}l\right|\le {\boldsymbol{\Omega}}_1\}$ is the frequency range of the wavelet bases. By \eqref{eq5}, let \\
\begin{align}
\beta\left(\eta \right)\triangleq& \mathop{sup}_{{| \boldsymbol{\omega} |}\in\left[1,2\right]}\sum_{ \begin{array}{c}
m\mathrm{\in }{\mathit{\Lambda}}_{re}\end{array}
}\left|\widehat{\phi }\left({| 2^{-m}\boldsymbol{\omega} |}\right)\right| \cdot\left|\widehat{\phi }\left({| 2^{-m}\boldsymbol{\omega} +\eta |}\right)\right| \nonumber\\
=&{\beta }_1\left(\eta \right)+{\beta }_2(\eta ).\nonumber
\end{align}

By the Cauchy–Schwarz inequality, we have
\begin{align}
&{\left({2\pi }\right)}^d\sum_{{{ \begin{array}{c}
l\mathrm{\in }{\textbf{Z}^d},m\mathrm{\in }{\mathit{\Lambda}}_{re} \end{array}
}}}{\int_{ \begin{array}{c}
{\boldsymbol{\omega}}\mathrm{\in }{\mathit{\Lambda}}_{fr}\end{array}
}}\overline{\widehat{\psi }\left(2^{-m}\boldsymbol{\omega} \right)}\widehat{\psi }(2^{-m}\boldsymbol{\omega} -{2\pi }l)\nonumber\\
&\cdot\overline{\widehat{f}(\boldsymbol{\omega} )}\widehat{f}(\boldsymbol{\omega} -\frac{\mathrm{2}\pi }{2^{-m}}l)d\boldsymbol{\omega} \nonumber\\
&\leqslant {\left({2\pi }\right)}^d\sum_{l\mathrm{\in }{\textbf{Z}^d}}{\left[\beta \left({2\pi }l\right)\beta \left(-{2\pi }l\right)\right]}^{\frac{1}{2}}{\|\left({\mathcal{P}}_{{\boldsymbol{\Omega}}_1}-{\mathcal{P}}_{{\boldsymbol{\Omega}}_0}\right)\widehat{f}\|}^2.\nonumber
\end{align}

For $m\ge m_0$, combining \eqref{eq16} we obtain
\begin{align}
&\sum_{l\mathrm{\in }{\textbf{Z}^d}}{\left[{\beta }_1\left({2\pi }l\right){\beta }_1\left(-{2\pi }l\right)\right]}^{\frac{1}{2}}\nonumber\\
&\le \sum_{l\mathrm{\in }{\textbf{Z}^d}}C_1{\left(1+{| {2\pi }l |}^2\right)}^{-(\gamma -\alpha )/2}\sum_{m=m_0}^{\infty}2^{-m\alpha }.\nonumber
\end{align}
Because $\gamma >\alpha +d$, $\sum_{l\mathrm{\in }{\textbf{Z}^d}}{\ }C_1{\left(1+{| {2\pi }l |}^2\right)}^{-(\gamma -\alpha )/2}$ is bounded. For ${\varepsilon }_1>0$, there exists some ${m}_0$ for which $\sum_{m=m_0}^{\infty}2^{-m\alpha }<{\varepsilon }_1$. Therefore, for some value of $C_3$, we have
\[\sum_{l\mathrm{\in }{\textbf{Z}^d}}C_1{\left(1+ | {{{2\pi }l} |}^2\right)}^{-(\gamma -\alpha )/2}\sum_{m=m_0}^{\infty}2^{-m\alpha }\le C_3{\varepsilon }_1.\]
For $m\le m_1$, combining \eqref{eq16} and specifying ${\varepsilon }_2>0$, we find that $\sum_{l\mathrm{\in }{\textbf{Z}^d}}C_2{\left(1+{| {2\pi }l |}^2\right)}^{-d(1+{\varepsilon }_2)/2}$ is bounded. When ${\varepsilon }_1>0$, there exists some $m_1$ for which $\sum_{m=-\infty}^{m_1}{\left(2^{-\delta (\gamma -\alpha )/2}\right)}^{-m}<{\varepsilon }_1$. Therefore, for some value of $C_4$, we have
\begin{align}
\sum_{l\mathrm{\in }{\textbf{Z}^d}}c_2{\left(1+{| {2\pi }l |}^2\right)}^{\frac{-d(1+{\varepsilon }_2)}{2}}\sum_{m=-\infty}^{m_1}{\left(2^{\frac{-\delta (\gamma -\alpha )}{2}}\right)}^{-m}
\le C_4{\varepsilon }_1.\nonumber
\end{align}
When $\varepsilon >0$, there exist some $m_0,\ m_1$ satisfying
\begin{align}
\label{eq19}
\sum_{{{\begin{array}{c}
n\mathrm{\in}{\textbf{Z}^d},m\mathrm{\in }{\mathit{\Lambda}}_{re} \end{array}
}}}{\left|\left\langle {\widehat{\psi }}_{mn},\left({\mathcal{P}}_{{\boldsymbol{\Omega}}_{\mathrm{1}}}\mathrm{-}{\mathcal{P}}_{{\boldsymbol{\Omega}}_0}\right)\widehat{f}\right\rangle \right|}^{\mathrm{2}}\le {\left({2\pi }\right)}^d{\varepsilon }^2{\|f\|}^2.
\end{align}

Now consider the fourth term on the right-hand side of Eq. \eqref{eq18}. Setting $m_0=1$ and $m_1=0$, we have
\[\sum_{{{{{ \begin{array}{c}
n\mathrm{\in }{\mathit{\Lambda}}_{tr}, m\mathrm{\in }\textbf{Z}
\end{array}
}}}}}{\left|\left\langle {\psi }_{mn},{\mathcal{Q}}_{\boldsymbol{T}}f\right\rangle \right|}^{\mathrm{2}}\]
\[\le {\left({2\pi }\right)}^d\sum_{{{{{ \begin{array}{c}
l\mathrm{\in }{\mathit{\Lambda}}_{tr}
 \end{array}
}}}}}{\left[\beta \left({2\pi }l\right)\beta \left(-{2\pi }l\right)\right]}^{\frac{1}{2}}{\cdot\|\left({\mathcal{Q}}_{\boldsymbol{T}}\right)f\|}^2.\]
As $m_0\ge 1$, $\sum_{m=m_0}^{\infty}2^{-m\alpha }\le \sum_{m=1}^{\infty}{\left(2^{\alpha}\right)}^{-m}=\frac{1}{{2^\alpha}-1}$. Thus, $\frac{1}{{2^\alpha}-1}$ is the upper bound of $\sum_{m=m_0}^{\infty}2^{-m\alpha }$. A similar upper bound exists for $\sum_{-\infty}^{m_1}{\left(2^{-\frac{\delta \left(\gamma -\alpha \right)}{2}}\right)}^{-m}$. Combining \eqref{eq16} and \eqref{eq17}, we obtain
\begin{align}
\beta \left(\eta \right)=&\mathop{sup}_{{| \boldsymbol{\omega} | }\in [1,2]}\ \sum_{m\mathrm{\in }\mathrm{z}}\left|\widehat{\phi }\left({| 2^{-m}\boldsymbol{\omega} | }\right)\right|\cdot \left|\widehat{\phi }\left({| 2^{-m}\boldsymbol{\omega} +\eta |}\right)\right| \nonumber \\
\le& C_{{\varepsilon }_2}{(1+{|\eta |}^2)}^{-d\frac{1+{\varepsilon }_2}{2}}, \nonumber
\end{align}
where $C_{{\varepsilon }_2}$ is a constant related to ${\varepsilon }_2$. We thus obtain
\begin{align}
&\sum_{l\mathrm{\in}{\mathit{\Lambda}}_{tr}}{\left[\beta\left({2\pi }l\right)\beta \left(-{2\pi }l\right)\right]}^{\frac{1}{2}}\nonumber\\
&\le C_{{\varepsilon }_2}{\left(\frac{1}{2\pi }\right)}^{d\left(1+{\varepsilon }_2\right)}\sum_{{{{{ \begin{array}{c}
l\mathrm{\in }{\mathit{\Lambda}}_{tr}
 \end{array}
}}}}}\nonumber\\
&{\left[{\left({\left(\frac{1}{2\pi }\right)}^2+{l_1}^2+{l_2}^2+\dots +{l_d}^2\right)}^d\right]}^{-\frac{1+{\varepsilon }_2}{2}}.\nonumber
\end{align}

By the inequality
\begin{align}
&{\left(C+{l_1}^2+{l_2}^2+\dots +{l_d}^2\right)}^d\nonumber\\
&\ge \left(C+{l_1}^2\right)\left(C+{l_2}^2\right)\dots (C+{l_d}^2),\nonumber
\end{align}
we get
\[\sum_{l\mathrm{\in }{\mathit{\Lambda}}_{tr}}{\left[\beta \left({2\pi }l\right)\beta \left(-{2\pi }l\right)\right]}^{\frac{1}{2}}\]
\[\le C_{{\varepsilon }_2}{\left(\frac{1}{2\pi }\right)}^{d\left(1+{\varepsilon }_2\right)}\sum_{{{{{ \begin{array}{c}
l\mathrm{\in }{\mathit{\Lambda}}_{tr}
 \end{array}
}}}}}{\ }\]
\[{\left[\frac{1}{\left({\left(\frac{1}{2\pi }\right)}^2+{l_1}^2\right)\left({\left(\frac{1}{2\pi }\right)}^2+{l_2}^2\right)\dots \left({\left(\frac{1}{2\pi }\right)}^2+{l_d}^2\right)}\right]}^{\frac{1+{\varepsilon }_2}{2}}.\]

Because $d$ is finite, for $\varepsilon >0$, there exists an enough large value $\boldsymbol{t}_{\varepsilon} = [t_1, t_2,\dots, t_d] $. For $\boldsymbol{t}_{\varepsilon} \preceq l$, it satisfies
\[C_{{\varepsilon }_2}{\left(\frac{1}{2\pi }\right)}^{d\left(1+{\varepsilon }_2\right)}\sum_{l\mathrm{\in }{\mathit{\Lambda}}_{tr}}{\ }\]
\[{\left[\frac{1}{\left({\left(\frac{1}{2\pi }\right)}^2+{l_1}^2\right)\left({\left(\frac{1}{2\pi }\right)}^2+{l_2}^2\right)\dots \left({\left(\frac{1}{2\pi }\right)}^2+{l_d}^2\right)}\right]}^{\frac{1+{\varepsilon }_2}{2}}\]
\[\le {\varepsilon }^2,\]
leading to
\begin{align}
\label{eq20}
    \sum_{{{{{ \begin{array}{c}
l\mathrm{\in }{\mathit{\Lambda}}_{tr},m\mathrm{\in }\mathrm{z} \end{array}
}}}}}{\left|\left\langle {\psi }_{mn},{\mathcal{Q}}_{\boldsymbol{T}}f\right\rangle \right|}^{\mathrm{2}}\le {\left(\frac{1}{2\pi }\right)}^d{\varepsilon }^2{\|f\|}^2.
\end{align}

Substituting \eqref{eq19} and \eqref{eq20} into \eqref{eq18}, we obtain
\[\left\|f-\sum_{\left(m,n\right)\in B_{\varepsilon }}{\left\langle {\psi }_{mn},f\right\rangle {\psi }_{mn}}\right\|\]
\[\le \|\left(\boldsymbol{\mathrm{1}}\mathrm{-}{\mathcal{P}}_{{\boldsymbol{\Omega}}_{\mathrm{1}}}\mathrm{+}{\mathcal{P}}_{{\boldsymbol{\Omega}}_0}\right)\widehat{f}\|\mathrm{+}\|\left(\boldsymbol{1}-{\mathcal{Q}}_{\boldsymbol{T}}\right)f\|+2{\left({2\pi }\right)}^{\frac{d}{2}}\varepsilon \| f\|,\]
which completes the proof. $\hfill\square$

\subsection{Proof of Corollary \ref{coro1}}\label{appendixC}
\emph{Proof:} From \eqref{energy_E_m} and \eqref{eq10}, 
\begin{align}
\label{eq10}
\widehat{E}_m=\displaystyle \sum_n{{\widehat{C}^2_{mn}\|{\psi}_{mn}(\boldsymbol{x})\|}^2},
\end{align}
we observe that ${\left\langle {\psi }_{m\mathrm{n}},f\right\rangle }^{\mathrm{2}}$ represents the contribution of ${\psi }_{mn}(\boldsymbol{x})$ to $\mathrm{\parallel }f{\mathrm{\parallel }}^{\mathrm{2}}$. As $\|f\|{\approx}\|({\mathcal{P}}_{{\boldsymbol{\Omega}}_1}-{\mathcal{P}}_{{\boldsymbol{\Omega}}_0} )\widehat{f}\|$ and $\|f\|{\approx}\|{\mathcal{Q}}_{\boldsymbol{T}} f\|$ in Assumption \ref{assum3}, we obtain
\[\|\left(\boldsymbol{1}-{\mathcal{P}}_{\boldsymbol{\Omega}_1}+{\mathcal{P}}_{\boldsymbol{\Omega }_0}\right)\widehat{f}\|+\|\left(\boldsymbol{1}-{\mathcal{Q}}_{\boldsymbol{T}}\right)f\|{\approx}0.\]

By Theorem \ref{thm1}, we then obtain
\begin{align}
&\left\|f-\sum_{\left(m,n\right)\in B_{\varepsilon }}{\left\langle{\psi }_{mn},f\right\rangle{\psi }_{mn}}\right\| \nonumber\\
\le& \|\left(\boldsymbol{\mathrm{1}}\mathrm{-}{\mathcal{P}}_{{\boldsymbol{\Omega}}_{\mathrm{1}}}\mathrm{+}{\mathcal{P}}_{{\boldsymbol{\Omega}}_0}\right)\widehat{f}\|\mathrm{+}\|\left(\boldsymbol{1}-{\mathcal{Q}}_{\boldsymbol{T}}\right)f\|+2{\left({2\pi }\right)}^{\frac{d}{2}}\varepsilon \|f\|\nonumber \\
{\approx}& 2{\left({2\pi }\right)}^{\frac{d}{2}}\varepsilon \|f\|.\nonumber
\end{align}

Because the wavelets are orthonormal, there exists a constant $D>1 $ satisfying
\begin{align*}
&\left\|\sum_{\left(m,n\right)\notin B_{\varepsilon }}{\left\langle{\psi }_{mn},f\right\rangle{\psi }_{mn}}\right\| \\
&=\sum_{\left(m,n\right)\notin B_{\varepsilon }}{\left|\left\langle{\psi }_{mn},f\right\rangle\right|}
< 2{D}{\left({2\pi }\right)}^{\frac{d}{2}}\varepsilon \| f\|.
\end{align*}

As $\varepsilon $ is sufficiently small, we have
\[\left|\langle{\psi }_{mn},f\rangle\right|\approx 0,\ \ \ \ \ \left(\left(m,n\right)\notin B_{\varepsilon }\right).  \]
The proof is completed. $\hfill\square$

\bibliographystyle{IEEEtran}
\input{CWNN_Arxiv_20250712.bbl}

\end{document}